%% file: main.tex
\PassOptionsToPackage{table}{xcolor}

\documentclass{article} 
\usepackage{iclr2027_conference}
\usepackage{times}

\input{math_commands.tex}

\usepackage{amsmath}
\usepackage{amsthm}
\usepackage{amsfonts}
\usepackage{amssymb}
\usepackage{caption}
\usepackage{capt-of}
\usepackage{needspace}
\usepackage{graphicx}
\usepackage{wrapfig}
\usepackage{subfigure}
\usepackage{booktabs}
\usepackage{multirow}
\usepackage{multicol}
\usepackage{threeparttable}

\usepackage{xcolor}
\usepackage{colortbl}

\definecolor{latrow}{RGB}{225,238,255}

\definecolor{hhhh}{RGB}{0,0,205}

\providecommand{\familyrule}{%
  \specialrule{0.35pt}{0.35ex}{0.35ex}%
}

\usepackage{algorithm}
\usepackage{algpseudocode}
\usepackage{arydshln}
\usepackage{ulem}
\usepackage{listings}
\usepackage[most]{tcolorbox}

\definecolor{promptborder}{HTML}{6F6F6F}
\definecolor{promptbg}{HTML}{FBFBFB}

\newtcblisting{promptbox}{
  enhanced,
  breakable,
  listing only,
  width=0.96\linewidth,
  center,
  colback=promptbg,
  colframe=promptborder,
  boxrule=0.55pt,
  arc=2.2mm,
  left=2.5mm,
  right=2.5mm,
  top=1.6mm,
  bottom=1.6mm,
  before skip=6pt,
  after skip=3pt,
  listing options={
    basicstyle=\ttfamily\scriptsize,
    breaklines=true,
    breakatwhitespace=true,
    breakindent=0pt,
    breakautoindent=false,
    columns=fullflexible,
    keepspaces=true,
    showstringspaces=false,
    literate={_}{{\_}}1
  }
}

\newcounter{prompt}

\newcommand{\promptcaption}[2]{%
  \refstepcounter{prompt}%
  \par\smallskip
  {\centering\small
  Prompt~\theprompt: #1%
  \label{#2}\par}%
  \medskip
}

\usepackage{url}
\usepackage{hyperref}

\hypersetup{
  colorlinks=true,
  allcolors=hhhh
}
\title{LaT: LLM-as-Trainer for Multi-Task Vehicle Routing Solvers}

\author{
Yang Wang$^{1}$,
Ya-Hui Jia$^{1}$\thanks{Corresponding author.} \,,
Wei-Neng Chen$^{2}$,
Yi Mei$^{3}$,
Wen Song$^{4}$,
Zhiguang Cao$^{5}$ \\
$^{1}$School of Future Technology, South China University of Technology, China\\
$^{2}$School of Computer Science and Engineering, South China University of Technology, China\\
$^{3}$School of Engineering and Computer Science, Victoria University of Wellington, New Zealand\\
$^{4}$Institute of Marine Science and Technology, Shandong University, China\\
$^{5}$School of Computing and Information Systems, Singapore Management University,
Singapore\\
\texttt{ftwangyang@mail.scut.edu.cn, jia.yahui@foxmail.com,}\\
\texttt{cwnraul634@aliyun.com, yi.mei@vuw.ac.nz,}\\
\texttt{wensong@email.sdu.edu.cn, zgcao@smu.edu.sg}
}

\iclrfinalcopy 
\begin{document}

\maketitle

\begin{abstract}
Multi-task neural solvers aim to handle multiple Vehicle Routing Problem (VRP) variants within a unified model, avoiding separate training for each constraint combination. However, VRP variants differ in optimization difficulty, while existing methods lack stage-wise feedback on their training status, making the model biased to some specific variants. Although meta-learning can support adaptive training, it typically requires bi-level optimization and additional gradient updates, increasing computational cost. To address this limitation, we propose \textbf{LLM-as-Trainer (LaT)}, a plug-and-play training paradigm that uses a pretrained large language model as an external trainer. LaT periodically analyzes cross-task validation metrics to generate a stage-wise guidance vector. This vector is combined with the current task’s constraint vector and injected into each encoder layer, providing the neural solver with additional training information during subsequent policy optimization. Experiments on 16 VRP variants show that LaT improves the solution quality of several state-of-the-art multi-task neural solvers on both trained and unseen variants, supporting the effectiveness and generality of the proposed training paradigm.
\end{abstract}

\section{Introduction}

Vehicle Routing Problems (VRPs) are a class of combinatorial optimization
problems with broad practical applications in logistics, transportation, and
supply chain management
\citep{toth2014vehicle,konstantakopoulos2022vehicle,vidal2020concise}. In
recent years, neural combinatorial optimization has emerged as a promising
paradigm for VRPs, in which constructive policies are trained by deep
reinforcement learning to generate feasible solutions efficiently
\citep{bengio2021machine,kool2018attention,kwon2020pomo}. Compared with
traditional heuristic solvers, neural solvers reduce the reliance on
hand-crafted rules and provide fast inference after training. However, early
neural solvers are usually designed and trained from scratch for a single VRP
variant
\citep{vinyals2015pointer,bello2017neural,nazari2018reinforcement,lu2020learning,falkner2020learning}.
Since real-world applications involve more than 60 VRP variants
\citep{vidal2020concise}, training a separate model for each constraint
combination is expensive and limits the scalability and practicality of neural
solvers in complex multi-constraint scenarios. Multi-task neural solvers
therefore aim to handle multiple VRP variants within a unified model, avoiding
separate training for each constraint combination.

Recent studies have developed unified models for solving multiple VRP variants
\citep{lin2024cross,berto2025routefinder,shield,moses,wang2026soft}. POMO-MTL represents VRP variants as combinations of basic constraints, enabling zero-shot generalization to unseen combinations \citep{mtl}. MVMoE uses a mixture-of-experts architecture to improve the model’s ability to handle diverse VRP variants \citep{zhou2024mvmoe}. CaDA uses
constraint prompts and dual attention to improve constraint awareness
\citep{li2024cada}. More recently, CCL dynamically updates node representations using the constraints most relevant at each construction step \citep{gui2026ccl}. PoMtVRS improves multi-task VRP solvers by refining decoder representations with a preference-gated block and ranking multiple solutions for each instance to train the policy to favor higher-quality ones \citep{meng2026pomt}.  Collectively, these methods have
improved the multi-task solving ability of neural solvers.

Despite this progress, joint training remains challenging because VRP variants differ in optimization difficulty. For example, multi-task solvers often show larger relative reference gaps on variants with time windows and backhauls than on capacity-only variants \citep{wang2024deep}. Because different variants share model parameters, updates from one variant can affect the model used for others. Existing multi-task optimization methods mainly coordinate these shared updates through training losses or gradients, without explicitly using stage-wise validation performance to guide joint training \citep{chen2018gradnorm,sener2018multi,yu2020gradient,wang2023efficient}. The optimization process does not explicitly respond to current cross-task performance differences, which may limit the overall solution quality of multi-task neural solvers. Meta-learning can introduce task-dependent adaptation during training \citep{finn2017model,hospedales2021survey}. However, it commonly relies on bi-level optimization with inner and outer loops, requiring additional gradient updates and increasing computational cost \citep{franceschi2018bilevel,nichol2018first}. This cost can become substantial for multi-task VRPs involving many tasks, complex constraint combinations, and long training runs. Recent LLM-assisted methods demonstrate that pretrained large language models
(LLMs) can improve the performance of neural solvers \citep{jiang2024unco,yang2024large,zhu2026refining}. For complex VRPs, DRoC decomposes
intricate constraints and retrieves constraint-specific knowledge to help LLMs
better exploit external solvers \citep{jiang2025droc}. Other studies use
LLM-derived representations to encode or align task semantics with spatial and
graph features \citep{malik2026llmaide,feng2026aligning}, or employ LLMs to
generate and evolve heuristics for guiding search and refining decoder
attention \citep{tran2025large,chi2026generalized}. However, these methods focus on
representing the optimization problem or modifying the solver heuristic. They
do not use the evolving validation status of multiple jointly trained VRP
variants to regulate subsequent policy optimization.


To address these limitations, we propose \textbf{LLM-as-Trainer (LaT)}, a
plug-and-play training paradigm that uses a pretrained LLM as an external
trainer for multi-task neural solvers. Unlike loss- or gradient-based task
coordination, LaT explicitly introduces cross-task validation information into
subsequent policy optimization. The LLM does not construct routes or directly
update model parameters. Instead, it periodically processes cross-task
validation metrics to generate a stage-wise guidance vector. For the task
currently being trained, this vector is combined with its constraint vector
and injected into each encoder layer, providing additional stage-wise training
information. This design avoids bi-level optimization and additional inner-loop
updates. The LLM remains frozen and is invoked only at predefined intervals,
introducing negligible additional training time. After training, the final
guidance vector is fixed and the LLM is removed, requiring no LLM API calls at
inference.

The main contributions of this paper are summarized as follows:
\begin{itemize}
    \item We propose \textbf{LaT}, a plug-and-play training paradigm that uses
    a pretrained LLM as an external trainer to introduce stage-wise task
    performance feedback into multi-task neural solvers. To the best of our
    knowledge, this is the first training paradigm that introduces an LLM as a
    trainer for multi-task vehicle routing solvers.
\item We design a lightweight mechanism that generates a stage-wise guidance vector from cross-task validation metrics. The vector is combined with the current task’s constraint vector and injected into each encoder layer, providing additional stage-wise training information during subsequent policy optimization.

\end{itemize}
We conduct systematic experiments on 16 VRP variants. The results
show that LaT improves the overall solution quality of multiple
state-of-the-art multi-task neural solvers across both trained and unseen
variants, demonstrating the effectiveness and generality of the proposed
training paradigm.

\section{Preliminaries}

\subsection{Multi-Task Vehicle Routing Problems}
We consider a family of \(16\) VRP variants built from
five elementary constraints, namely capacity (C), open route (O), backhaul (B),
route-length limit (L), and time window (TW). These constraints form the set
\(\mathcal{M}=\{\mathrm{C},\mathrm{O},\mathrm{B},\mathrm{L},\mathrm{TW}\}\), and a
single constraint is indexed by \(m\in\mathcal{M}\). Each task
\(k\in\{1,\ldots,16\}\) is described by a binary constraint vector:
\begin{equation}
    \mathbf{z}_k=
    \bigl[z_k^{\mathrm C},z_k^{\mathrm O},z_k^{\mathrm B},z_k^{\mathrm L},z_k^{\mathrm{TW}}\bigr]
    \in\{0,1\}^{5},
    \qquad
    z_k^{m}=
    \begin{cases}
        1, & \text{constraint } m \text{ is active in task } k,\\
        0, & \text{otherwise},
    \end{cases}
    \label{eq:zvec}
\end{equation}
whose entries follow the fixed order of \(\mathcal{M}\). The capacity constraint
is active in every task, and the other variants activate different subsets of
\(\{\mathrm{O},\mathrm{B},\mathrm{L},\mathrm{TW}\}\). The complete list of the
\(16\) tasks is given in Appendix~\ref{app:mtvrp-details}.

A feasible solution \(\tau\) is a set of routes that start from the depot and
visit every customer once while respecting the active constraints of
\(\mathbf{z}_k\). Let \(\mathcal{F}_k\) denote the feasible set of task \(k\).
The objective minimizes the total travel cost:
\begin{equation}
   \tau_k^{\star}=\arg\min_{\tau\in\mathcal{F}_k} c(\tau),\qquad c(\tau)=\sum_{\rho\in\tau}\sum_{j=1}^{|\rho|-1} d_{\rho_j,\rho_{j+1}},
    \label{eq:obj}
\end{equation}
where \(\rho\) is a single route and the final customer-to-depot edge is omitted
when the open-route constraint is active. A multi-task neural solver approximates
\(\tau_k^{\star}\) for all \(16\) tasks with one shared set of parameters.

\subsection{Autoregressive Construction and Policy Optimization}
A policy \(\pi_\theta\) constructs a solution step by step. At step \(t\), the
state \(s_t\) summarizes the partial route, the feasible nodes, and the dynamic
route attributes \(\mathbf{d}_t=[Q_t^{\mathrm{rem}},\vartheta_t,D_t,o_t]\), which collect the remaining
capacity, the current time, the accumulated route length, and the open-route
indicator. The action \(a_t\) selects the next node from those that remain
feasible under \(\mathbf{z}_k\), and the probability of a complete solution
factorizes over the \(T_\tau\) construction steps:
\begin{equation}
    \pi_\theta(\tau\mid G,\mathbf{z}_k)
    =\prod_{t=1}^{T_\tau}\pi_\theta\bigl(a_t\mid s_t,G,\mathbf{z}_k\bigr).
    \label{eq:policy}
\end{equation}
The solver is a Transformer encoder-decoder. The encoder runs once per instance
and produces the node embeddings \(\mathbf{h}_i^{\ell}\) at layer
\(\ell\in\{1,\ldots,L_{\mathrm{enc}}\}\), and the decoder reuses the final-layer embeddings at
every step. The policy is trained by a POMO-style policy gradient that rolls out
\(N\) trajectories from \(N\) distinct starting nodes and uses their mean return
as a shared baseline \citep{kwon2020pomo}:
\begin{equation}
    \nabla_\theta\mathcal{L}(\theta)
    \approx-\frac{1}{N}\sum_{p=1}^{N}
    \bigl(\mathcal{R}(\tau^{p})-\overline{\mathcal{R}}(G)\bigr)\,
    \nabla_\theta\log\pi_\theta\bigl(\tau^{p}\mid G,\mathbf{z}_k\bigr),
    \qquad
    \overline{\mathcal{R}}(G)=\frac{1}{N}\sum_{p=1}^{N}\mathcal{R}(\tau^{p}),
    \label{eq:pg}
\end{equation}
where \(p\) indexes the POMO trajectories and \(\mathcal{R}(\tau^{p})=-c(\tau^{p})\). 

\section{Methodology}
\label{sec_method}

\subsection{Overview}

\begin{figure}[t]
    \centering
    \includegraphics[width=0.99\linewidth]{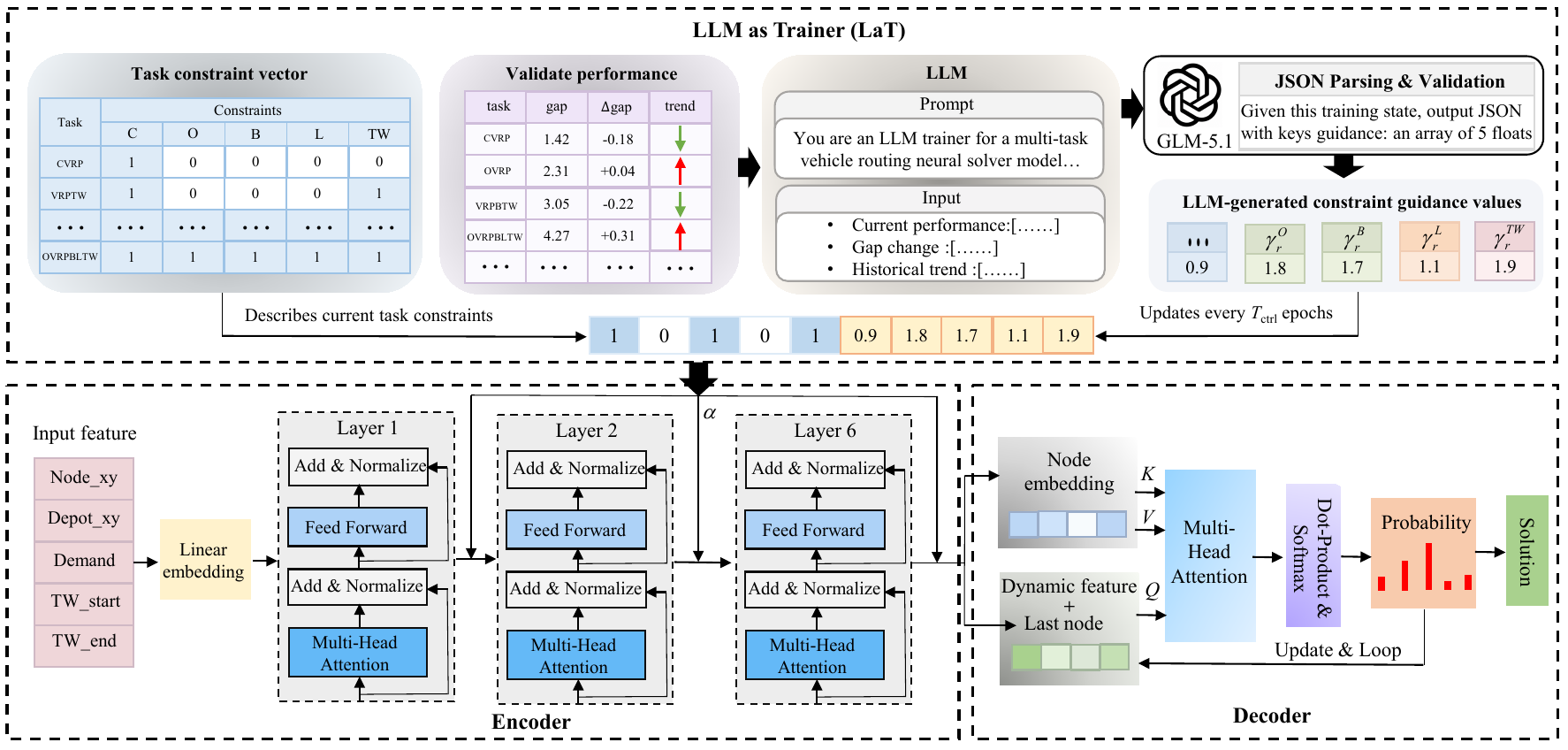}
    \caption{Overview of LaT. At predefined intervals, relative reference gaps
     from all validation tasks are organized into a
    structured training state. A pretrained LLM processes this state to generate a
    five-dimensional guidance vector corresponding to the five elementary
    constraints. During subsequent optimization, the vector is combined with
    the current task's constraint vector and injected after each encoder layer
    through a lightweight side branch. After training, the final guidance
    vector is fixed and the LLM is removed.}
    \label{fig_lat_overview}
\end{figure}

As shown in Fig.~\ref{fig_lat_overview}, LaT extends the training process
without changing the original autoregressive policy, backbone
encoder--decoder architecture, or reinforcement learning objective. At
predefined intervals, the solver is evaluated on all \(16\) validation tasks.
Their relative reference gaps  are organized into
a structured training state and provided to a pretrained LLM. The LLM generates a
five-dimensional guidance vector, with one value corresponding to each
elementary constraint.

During subsequent optimization, the latest guidance vector is combined with
the constraint vector of the current task and injected after each encoder layer
through a lightweight side branch. The decoder uses the resulting encoder
embeddings as keys and values, allowing the guidance information to affect
solution construction without modifying the decoder. The guidance vector is
updated at the next predefined interval.

The LLM receives no gradients and is invoked only periodically, introducing
negligible additional training time. After training, the final guidance vector
is fixed and the LLM is removed. The fixed vector and learned side branches
remain in the solver, requiring no LLM API calls and introducing negligible
computational overhead at inference.

\subsection{Guidance Generation}
\label{sec:guidance}
This subsection explains how LaT constructs a stage-wise training state from
cross-task validation results and uses it to generate a 
guidance vector. Each task \(k\) retains the binary constraint vector
\(\mathbf{z}_k\) defined in Eq.~\ref{eq:zvec}, which remains fixed throughout
training.

\paragraph{Cross-task validation.}
LaT performs cross-task validation every \(T_{\mathrm{ctrl}}\) training
epochs. At each control round, the current policy is evaluated on a fixed
validation set for each of the \(16\) tasks. All validation instances are
generated before training and are disjoint from the training and test
instances. The validation
objective of task \(k\) at control round \(r\) is:
\begin{equation}
    y_{r,k}
    =
    \frac{1}{|\mathcal{D}_k|}
    \sum_{G\in\mathcal{D}_k}
    c\bigl(\tau_{\theta}(G)\bigr),
    \label{eq:val}
\end{equation}
where \(\tau_{\theta}(G)\) is the solution constructed under
\(\mathbf{z}_k\). The validation results are subsequently used to construct
the training state and generate \(\boldsymbol{\gamma}_r\), which is applied
from the next training epoch.

\paragraph{Reference objective and relative reference gap.}
For each validation instance \(G\in\mathcal{D}_k\), a reference
solution \(\tau^{\mathrm{ref}}(G)\) is computed offline. For CVRP and
VRPTW, the reference solutions are generated by
HGS~\citep{vidal2022hybrid}. For the remaining \(14\) variants, they
are generated by OR-Tools~\citep{ortools_routing} with a time limit of
\(200\) seconds per instance. The mean reference objective of task
\(k\) is defined as:
\begin{equation}
    y_k^{\mathrm{ref}}
    =
    \frac{1}{|\mathcal{D}_k|}
    \sum_{G\in\mathcal{D}_k}
    c\bigl(\tau^{\mathrm{ref}}(G)\bigr),
    \label{eq:ref}
\end{equation}
where \(y_k^{\mathrm{ref}}\) serves as a fixed reference for measuring
the validation performance of the neural solver.

The relative reference gap of task \(k\) at control round \(r\), and
its change from the previous round, are defined as:
\begin{equation}
    \eta_{r,k}
    =
    \frac{y_{r,k}-y_k^{\mathrm{ref}}}
         {y_k^{\mathrm{ref}}},
    \qquad
    \Delta\eta_{r,k}
    =
    \eta_{r,k}-\eta_{r-1,k},
    \qquad
    \Delta\eta_{1,k}=0.
    \label{eq:gap}
\end{equation}
A larger \(\eta_{r,k}\) corresponds to a larger observed gap from the fixed
reference objective. A positive \(\Delta\eta_{r,k}\) indicates that this gap
has increased since the previous control round.

\paragraph{Training state.}
For each task \(k\), LaT retains the \(J\) most recent validation records, which
consist of the current record \((y_{r,k},\eta_{r,k})\) and the \(J-1\) records
preceding it. The preceding records are collected as
$\mathcal{H}_{r,k}=\left\{(y_{r',k},\eta_{r',k})\right\}_{r'=\max(1,r-J+1)}^{r-1}$,
which is empty at the first control round, i.e. $\mathcal{H}_{1,k}=\varnothing$.
 The records of all tasks, together with the initial guidance and the guidance used before the current update, are organized into the training state:
\begin{equation}
    \mathcal{I}_r
    =
    \left(
    \boldsymbol{\gamma}_0,
    \boldsymbol{\gamma}_{r-1},
    \left\{
    \left(
    k,\mathbf{z}_k,y_{r,k},\eta_{r,k},
    \Delta\eta_{r,k},\mathcal{H}_{r,k}
    \right)
    \right\}_{k=1}^{16}
    \right),
    \label{eq:state}
\end{equation}
where \(\boldsymbol{\gamma}_0\) is the initial guidance and
\(\boldsymbol{\gamma}_{r-1}\) is the guidance applied during the training
interval preceding control round \(r\). 

\paragraph{Guidance vector.}
LaT writes the training state \(\mathcal{I}_r\) into a fixed prompt
template \(\mathcal{P}\) and queries the pretrained LLM to generate a
five-dimensional guidance vector:
\begin{equation}
    \boldsymbol{\gamma}_r
    =
    \bigl[
    \gamma_r^{\mathrm C},
    \gamma_r^{\mathrm O},
    \gamma_r^{\mathrm B},
    \gamma_r^{\mathrm L},
    \gamma_r^{\mathrm{TW}}
    \bigr]
    =
    \mathrm{LLM}\bigl(\mathcal{I}_r;\mathcal{P}\bigr)
    \in\mathbb{R}^{5},
    \label{eq:guidance}
\end{equation}
where the entries follow the same order as those of
\(\mathbf{z}_k\). Each element of \(\boldsymbol{\gamma}_r\) corresponds to one of the five
elementary constraints. The LLM generates this vector from the validation
results of all \(16\) tasks. The same guidance
vector is used for all tasks until the next control round. When training task
\(k\), \(\boldsymbol{\gamma}_r\) is concatenated with
\(\mathbf{z}_k\), which indicates the constraints active in that task. The
combined vector is then provided to each encoder side branch. The complete
prompt is presented in Appendix~\ref{app:prompts}.

\subsection{Encoder}
\label{sec:encoder}

LaT injects the guidance vector into each encoder layer through a lightweight
side branch scaled by a learnable coefficient. During the training interval
following control round \(r\), the side-branch input for task \(k\) is formed
by concatenating its constraint vector with the latest guidance vector:
\begin{equation}
    \mathbf{u}_{r,k}
    =
    [\mathbf{z}_k,\boldsymbol{\gamma}_r]
    \in\mathbb{R}^{10}.
    \label{eq:side-input}
\end{equation}
The concatenated vector \(\mathbf{u}_{r,k}\) is provided to the side branch
at every encoder layer. 

The encoder maps the static feature
\(\mathbf{f}_i=[\mathbf{x}_i,q_i,e_i,l_i]\) of node \(v_i\), including the
depot \(v_0\), to an initial embedding through a linear projection. Features
inactive for the current task are padded to zero. We initialize the encoder
recurrence as:
\begin{equation}
    \mathbf{h}_i^{0}
    =
    W_{\mathrm{in}}\mathbf{f}_i,
    \qquad
    \widetilde{\mathbf{h}}_i^{0}
    =
    \mathbf{h}_i^{0},
    \qquad
    i=0,\ldots,N.
    \label{eq:initial-embedding}
\end{equation}

At encoder layer \(\ell\), the backbone layer receives the outputs of the
preceding LaT layer and produces the backbone-layer output
\(\mathbf{h}_i^\ell\):
\begin{equation}
    \mathbf{h}_i^\ell
    =
    \mathcal{B}_\ell
    \left(
        \left\{
        \widetilde{\mathbf{h}}_j^{\ell-1}
        \right\}_{j=0}^{N}
    \right),
    \qquad
    \ell=1,\ldots,L_{\mathrm{enc}},
    \label{eq:backbone-layer}
\end{equation}
where \(\mathcal{B}_\ell\) denotes the multi-head attention and feed-forward
computations of backbone layer \(\ell\), as detailed in
Appendix~\ref{app:backbone}. The side branch then combines
\(\mathbf{h}_i^\ell\) with \(\mathbf{u}_{r,k}\):
\begin{equation}
    \psi_\ell
    \bigl(
        \mathbf{h}_i^\ell,
        \mathbf{u}_{r,k}
    \bigr)
    =
    W_{\mathrm{side}}^\ell
    \mathrm{ReLU}
    \left(
        W_h^\ell\mathbf{h}_i^\ell
        +
        W_u^\ell\mathbf{u}_{r,k}
    \right),
    \label{eq:branch}
\end{equation}
where
\(W_h^\ell\in
\mathbb{R}^{d_{\mathrm{side}}\times d_{\mathrm{model}}}\) and
\(W_u^\ell\in
\mathbb{R}^{d_{\mathrm{side}}\times 10}\) map
\(\mathbf{h}_i^\ell\) and \(\mathbf{u}_{r,k}\) into a space of dimension
\(d_{\mathrm{side}}\), respectively. The output projection
\(W_{\mathrm{side}}^\ell\in
\mathbb{R}^{d_{\mathrm{model}}\times d_{\mathrm{side}}}\) maps the side-branch
output back to the model dimension.

The side-branch output is added to the backbone-layer output through the
learnable coefficient \(\alpha\):
\begin{equation}
    \widetilde{\mathbf{h}}_i^\ell
    =
    \mathbf{h}_i^\ell
    +
    \alpha\,
    \psi_\ell
    \bigl(
        \mathbf{h}_i^\ell,
        \mathbf{u}_{r,k}
    \bigr).
    \label{eq:inject}
\end{equation}
A single learnable coefficient \(\alpha\) is shared across all
encoder layers and optimized jointly with the solver parameters. The resulting \(\widetilde{\mathbf{h}}_i^\ell\) is used as the input to
backbone layer \(\ell+1\), as specified in
Eq.~\ref{eq:backbone-layer}. Consequently, the output of a side branch can
affect all subsequent encoder layers.

\subsection{Decoder}
\label{sec:decoder}

The decoder retains its original architecture and uses the final encoder
embeddings \(\widetilde{\mathbf{h}}_i^{L_{\mathrm{enc}}}\) as keys and values.
These embeddings are also used to construct the decoder context and the keys
of the final compatibility layer. Therefore, the information introduced by
the encoder side branches is passed to the node-selection policy without
additional decoder modules.

At each decoding step, the context consists of the embedding of the last
selected node and the dynamic state
\(\mathbf{d}_t=[Q_t^{\mathrm{rem}},\vartheta_t,D_t,o_t]\).
The decoder computes the selection probabilities using multi-head attention
and a single-head compatibility layer with \(\tanh\) clipping. Nodes that
violate the active constraints in \(\mathbf{z}_k\) are masked. The complete
decoder computations are provided in Appendix~\ref{app:backbone}.

\section{Experiments}
\subsection{Experimental Setup}
Following \citep{zhou2024mvmoe}, we evaluate LaT on \(16\) VRP variants
formed from five elementary constraints. All experiments are implemented in
PyTorch and conducted on a single NVIDIA RTX A6000 GPU. We use GLM-5.1 \citep{glm5team2026glm5},
provided through the Zhipu AI BigModel Open Platform, as the external trainer.

\paragraph{Training Settings.}
We follow the problem generation and general training settings of
\citep{mtl,zhou2024mvmoe}. For each LaT variant, we retain the architecture
and solver-specific settings of its corresponding backbone. Separate models
are trained for problem sizes of 50 and 100 nodes. For each problem size, the
models are jointly trained on six variants: CVRP, OVRP, VRPB, VRPL, VRPTW,
and OVRPTW. They are then evaluated zero-shot on the remaining ten variants
of the same size. Each model is trained for \(5{,}000\) epochs with a batch
size of 128 and a total of \(20{,}000\) training instances per epoch, resulting
in \(100\) million training instances. At the first control round,
\(\boldsymbol{\gamma}_0=\mathbf{1}\). The guidance vector is updated every
\(T_{\mathrm{ctrl}}=500\) epochs, and the latest \(J=5\) validation records
for each task are retained, i.e. the current record and the four preceding ones. A sensitivity analysis of the guidance-update interval is provided in
Appendix~\ref{app:update-interval-sensitivity}.

\paragraph{Comparison Methods.} We compare LaT with representative traditional and neural solvers. The
traditional solvers include HGS \citep{vidal2022hybrid}, LKH3
\citep{helsgaun}, and OR-Tools \citep{ortools_routing}, whose configurations
follow \citep{kool2018attention}. The neural solvers include the single-task
solver POMO \citep{kwon2020pomo} and the multi-task solvers POMO-MTL
\citep{mtl}, MVMoE, MVMoE-L \citep{zhou2024mvmoe}, RF-POMO
\citep{berto2025routefinder}, CaDA \citep{li2024cada}, ReLD
\citep{huang2025rethinking}, SPSM \citep{wang2026soft}, and CCL \citep{gui2026ccl}. To evaluate whether
LaT can be applied to different neural solvers, we integrate it with ReLD and
CaDA and denote the resulting models as \textbf{LaT-ReLD} and
\textbf{LaT-CaDA}, respectively.  Additional experimental results are provided in
Appendix~\ref{app:additional-experiments}.


\paragraph{Inference Settings.} All neural solvers, including the LaT variants, use the eight-fold instance
augmentation introduced in \citep{kwon2020pomo}. Each solver is evaluated on
\(1{,}000\) test instances for every VRP variant and problem size. We report
the average objective value, the relative gap to the best-performing
non-learning solver, and the total inference time required to solve the
complete test set.

\subsection{Experimental Results}

\subsubsection{Evaluation of Performance on Trained VRPs}

To evaluate whether LaT improves solution quality on the VRP variants used
during training and can be applied to different neural solvers, we compare
LaT-ReLD and LaT-CaDA with their corresponding backbones. As shown in
Table~\ref{tab:1}, both LaT variants achieve lower gaps than their backbones
across all six trained VRPs and both problem sizes. LaT reduces the mean gap
of ReLD from \(1.923\%\) to \(1.756\%\) and that of CaDA from \(1.810\%\) to
\(1.669\%\). The improvements are particularly clear on variants involving
time windows. For example, LaT-ReLD reduces the gap on 100-node OVRPTW from
\(3.214\%\) to \(2.737\%\), while LaT-CaDA reduces the gap on 50-node OVRPTW
from \(2.482\%\) to \(2.232\%\).

Among all multi-task neural solvers, a LaT variant achieves the lowest gap in
\(10\) of the \(12\) settings, while CCL performs best in the remaining two
settings. When averaged over all settings, LaT-CaDA achieves the lowest mean
gap of \(1.669\%\), showing strong overall performance across the trained
variants. The consistent improvements obtained with both ReLD and CaDA also
show that LaT can be integrated with different neural solvers. After training,
the LLM is removed, while the final guidance vector and learned encoder side
branches remain in the deployed solver. The LaT variants require \(3\)--\(4\)
seconds for 50-node instances and \(10\)--\(17\) seconds for 100-node
instances, which is close to the inference time of their corresponding
backbones. This indicates that LaT improves solution quality with little
additional inference cost.

\subsubsection{Evaluation of Performance on Unseen VRPs}

To evaluate the zero-shot generalization of LaT, we test LaT-ReLD and
LaT-CaDA on the remaining ten unseen VRP variants without further training.
As shown in Table~\ref{tab:2}, LaT-CaDA outperforms CaDA in all \(20\)
settings, reducing the mean gap from \(5.490\%\) to \(5.067\%\). Its largest
improvement is obtained on 100-node OVRPBL, where the gap decreases from
\(7.326\%\) to \(5.813\%\). LaT-ReLD also reduces the mean gap from
\(5.328\%\) to \(5.240\%\), although its improvements are less consistent
across individual variants.

Across the \(20\) unseen settings, a LaT variant achieves the lowest gap in
\(14\) settings, while CCL performs best in the remaining six. LaT-CaDA
achieves the lowest overall mean gap among all multi-task neural solvers,
showing that its improvements extend to constraint combinations not used
during training. The LaT variants also retain inference times close to their
backbones, requiring \(2\)--\(4\) seconds for 50-node instances and
\(10\)--\(18\) seconds for 100-node instances. These results show that LaT
provides strong overall zero-shot performance without sacrificing inference
efficiency. The different improvements obtained with ReLD and CaDA also
indicate that the benefit of LaT depends partly on the selected backbone.
Further zero-shot evaluations on CVRPLIB Set-X and Set-Solomon instances are
reported in Appendix~\ref{app:benchmark-results}.

\begin{table*}[!t]
  \centering
  \caption{Performance on 1K test instances from the six training VRP
  variants. Non-learning solvers and the single-task POMO are above the dashed line,
  while the multi-task neural solvers are below. The best-performing
  non-learning solver is used to calculate the gap. The best results
  among the multi-task neural solvers are highlighted in \textbf{bold}.}
  \label{tab:1}

  \begin{small}
  \setlength{\tabcolsep}{2.8pt}
  \renewcommand{\arraystretch}{1.05}

  \resizebox{\textwidth}{!}{%
  \begin{tabular}{
    @{}llcccccc
    @{\hspace{8pt}}
    llcccccc@{}
  }
    \toprule
    \multicolumn{2}{c}{\multirow{2}{*}{Method}}
    & \multicolumn{3}{c}{$N=50$}
    & \multicolumn{3}{c}{$N=100$}
    & \multicolumn{2}{c}{\multirow{2}{*}{Method}}
    & \multicolumn{3}{c}{$N=50$}
    & \multicolumn{3}{c}{$N=100$}
    \\
    \cmidrule(lr){3-5}
    \cmidrule(lr){6-8}
    \cmidrule(lr){11-13}
    \cmidrule(lr){14-16}
    & & Obj. & Gap & Time
    & Obj. & Gap & Time
    && & Obj. & Gap & Time
    & Obj. & Gap & Time
    \\
    \midrule

    \multirow{15}{*}{\rotatebox{90}{CVRP}}
    & HGS
    & 10.334 & 0.000\% & 4.6m
    & 15.504 & 0.000\% & 9.1m
    & \multirow{15}{*}{\rotatebox{90}{VRPTW}}
    & HGS
    & 14.509 & 0.000\% & 8.4m
    & 24.339 & 0.000\% & 19.6m
    \\
    & LKH3
    & 10.346 & 0.115\% & 9.9m
    & 15.590 & 0.556\% & 18.0m
    & & LKH3
    & 14.607 & 0.664\% & 5.5m
    & 24.721 & 1.584\% & 7.8m
    \\
    & OR-Tools
    & 10.540 & 1.962\% & 10.4m
    & 16.381 & 5.652\% & 20.8m
    & & OR-Tools
    & 14.915 & 2.694\% & 10.4m
    & 25.894 & 6.297\% & 20.8m
    \\
    & OR-Tools ($\times 10$)
    & 10.418 & 0.788\% & 1.7h
    & 15.935 & 2.751\% & 3.5h
    & & OR-Tools ($\times 10$)
    & 14.665 & 1.011\% & 1.7h
    & 25.212 & 3.482\% & 3.5h
    \\
    & POMO
    & 10.418 & 0.806\% & 3s
    & 15.734 & 1.488\% & 9s
    & & POMO
    & 14.940 & 2.971\% & 3s
    & 25.367 & 4.307\% & 12s
    \\
    \cdashline{2-8}\cdashline{10-16}
    & POMO-MTL
    & 10.437 & 0.987\% & 3s
    & 15.790 & 1.846\% & 9s
    & & POMO-MTL
    & 15.032 & 3.605\% & 3s
    & 25.610 & 5.313\% & 12s
    \\
    & MVMoE
    & 10.428 & 0.896\% & 4s
    & 15.760 & 1.653\% & 12s
    & & MVMoE
    & 14.999 & 3.377\% & 4s
    & 25.512 & 4.903\% & 15s
    \\
    & MVMoE-L
    & 10.434 & 0.955\% & 4s
    & 15.771 & 1.728\% & 11s
    & & MVMoE-L
    & 15.013 & 3.474\% & 4s
    & 25.519 & 4.927\% & 14s
    \\
    & RF-POMO
    & 10.431 & 0.926\% & 3s
    & 15.761 & 1.661\% & 9s
    & & RF-POMO
    & 14.997 & 3.363\% & 3s
    & 25.505 & 4.875\% & 13s
    \\
    & SPSM
    & 10.425 & 0.873\% & 4s
    & 15.740 & 1.531\% & 14s
    & & SPSM
    & 14.983 & 3.267\% & 4s
    & 25.461 & 4.694\% & 17s
    \\
    & ReLD
    & 10.420 & 0.814\% & 4s
    & 15.730 & 1.462\% & 10s
    & & ReLD
    & 14.977 & 3.226\% & 4s
    & 25.429 & 4.554\% & 13s
    \\
    & CaDA
    & 10.430 & 0.919\% & 4s
    & 15.735 & 1.500\% & 14s
    & & CaDA
    & 14.966 & 3.150\% & 4s
    & 25.365 & 4.289\% & 17s
    \\
    & CCL
    & 10.430 & 0.928\% & 6s
    & 15.755 & 1.624\% & 21s
    & & CCL
    & \textbf{14.925} & \textbf{2.868\%} & 7s
    & 25.358 & 4.254\% & 24s
    \\
    \rowcolor{latrow}
    & LaT-ReLD
    & \textbf{10.419} & \textbf{0.810\%} & 4s
    & \textbf{15.718} & \textbf{1.385\%} & 10s
    & & LaT-ReLD
    & 14.947 & 3.019\% & 4s
    & 25.380 & 4.343\% & 13s
    \\
    \rowcolor{latrow}
    & LaT-CaDA
    & 10.420 & 0.823\% & 4s
    & 15.723 & 1.411\% & 14s
    & & LaT-CaDA
    &14.929 & 2.903\% & 4s
    & \textbf{25.341} & \textbf{4.179\%} & 17s
    \\
    \midrule

    \multirow{14}{*}{\rotatebox{90}{OVRP}}
    & LKH3
    & 6.511 & 0.198\% & 4.5m
    & 9.828 & 0.000\% & 5.3m
    & \multirow{14}{*}{\rotatebox{90}{VRPL}}
    & LKH3
    & 10.571 & 0.790\% & 7.8m
    & 15.771 & 0.000\% & 16.0m
    \\
    & OR-Tools
    & 6.531 & 0.495\% & 10.4m
    & 10.010 & 1.806\% & 20.8m
    & & OR-Tools
    & 10.677 & 1.746\% & 10.4m
    & 16.496 & 4.587\% & 20.8m
    \\
    & OR-Tools ($\times 10$)
    & 6.498 & 0.000\% & 1.7h
    & 9.842 & 0.122\% & 3.5h
    & & OR-Tools ($\times 10$)
    & 10.495 & 0.000\% & 1.7h
    & 16.004 & 1.444\% & 3.5h
    \\
    & POMO
    & 6.609 & 1.685\% & 2s
    & 10.044 & 2.192\% & 9s
    & & POMO
    & 10.491 & -0.038\% & 2s
    & 15.785 & 0.093\% & 10s
    \\
    \cdashline{2-8}\cdashline{10-16}
    & POMO-MTL
    & 6.671 & 2.634\% & 2s
    & 10.169 & 3.458\% & 9s
    & & POMO-MTL
    & 10.513 & 0.172\% & 2s
    & 15.846 & 0.479\% & 10s
    \\
    & MVMoE
    & 6.655 & 2.402\% & 3s
    & 10.138 & 3.136\% & 12s
    & & MVMoE
    & 10.501 & 0.057\% & 3s
    & 15.812 & 0.261\% & 14s
    \\
    & MVMoE-L
    & 6.665 & 2.548\% & 3s
    & 10.145 & 3.214\% & 11s
    & & MVMoE-L
    & 10.506 & 0.105\% & 3s
    & 15.821 & 0.323\% & 12s
    \\
    & RF-POMO
    & 6.652 & 2.351\% & 2s
    & 10.135 & 3.109\% & 9s
    & & RF-POMO
    & 10.505 & 0.095\% & 2s
    & 15.817 & 0.299\% & 10s
    \\
    & SPSM
    & 6.644 & 2.223\% & 3s
    & 10.101 & 2.770\% & 14s
    & & SPSM
    & 10.499 & 0.038\% & 3s
    & 15.796 & 0.165\% & 15s
    \\
    & ReLD
    & 6.642 & 2.202\% & 3s
    & 10.066 & 2.412\% & 11s
    & & ReLD
    & 10.497 & 0.019\% & 3s
    & 15.783 & 0.082\% & 12s
    \\
    & CaDA
    & 6.623 & 1.906\% & 3s
    & 10.047 & 2.227\% & 13s
    & & CaDA
    & 10.503 & 0.076\% & 3s
    & 15.788 & 0.119\% & 14s
    \\
    & CCL
    & 6.618 & 1.827\% & 6s
    & 10.084 & 2.597\% & 21s
    & & CCL
    & 10.507 & 0.114\% & 6s
    & 15.812 & 0.272\% & 20s
    \\
    \rowcolor{latrow}
    & LaT-ReLD
    & 6.624 & 1.919\% & 3s
    & 10.042 & 2.168\% & 11s
    & & LaT-ReLD
    & 10.495 & 0.000\% & 3s
    & \textbf{15.774} & \textbf{0.025\%} & 12s
    \\
    \rowcolor{latrow}
    & LaT-CaDA
    & \textbf{6.607} & \textbf{1.664\%} & 3s
    & \textbf{10.039} & \textbf{2.131\%} & 13s
    & & LaT-CaDA
    & \textbf{10.494} & \textbf{-0.010\%} & 3s
    & 15.777 & 0.047\% & 14s
    \\
    \midrule

    \multirow{13}{*}{\rotatebox{90}{VRPB}}
    & OR-Tools
    & 8.127 & 0.989\% & 10.4m
    & 12.185 & 2.594\% & 20.8m
    & \multirow{13}{*}{\rotatebox{90}{OVRPTW}}
    & OR-Tools
    & 8.737 & 0.592\% & 10.4m
    & 14.635 & 1.756\% & 20.8m
    \\
    & OR-Tools ($\times 10$)
    & 8.046 & 0.000\% & 1.7h
    & 11.878 & 0.000\% & 3.5h
    & & OR-Tools ($\times 10$)
    & 8.683 & 0.000\% & 1.7h
    & 14.380 & 0.000\% & 3.5h
    \\
    & POMO
    & 8.149 & 1.276\% & 2s
    & 11.993 & 0.968\% & 8s
    & & POMO
    & 8.891 & 2.377\% & 3s
    & 14.728 & 2.467\% & 12s
    \\
    \cdashline{2-8}\cdashline{10-16}
    & POMO-MTL
    & 8.182 & 1.684\% & 2s
    & 12.072 & 1.633\% & 8s
    & & POMO-MTL
    & 8.987 & 3.470\% & 3s
    & 15.008 & 4.367\% & 12s
    \\
    & MVMoE
    & 8.170 & 1.540\% & 3s
    & 12.027 & 1.254\% & 10s
    & & MVMoE
    & 8.964 & 3.210\% & 4s
    & 14.927 & 3.852\% & 15s
    \\
    & MVMoE-L
    & 8.176 & 1.605\% & 3s
    & 12.036 & 1.330\% & 10s
    & & MVMoE-L
    & 8.974 & 3.322\% & 4s
    & 14.940 & 3.941\% & 14s
    \\
    & RF-POMO
    & 8.169 & 1.520\% & 2s
    & 12.029 & 1.271\% & 8s
    & & RF-POMO
    & 8.950 & 3.060\% & 3s
    & 14.916 & 3.777\% & 12s
    \\
    & SPSM
    & 8.163 & 1.444\% & 3s
    & 12.007 & 1.086\% & 12s
    & & SPSM
    & 8.936 & 2.899\% & 4s
    & 14.872 & 3.470\% & 17s
    \\
    & ReLD
    & 8.155 & 1.352\% & 3s
    & 11.987 & 0.918\% & 10s
    & & ReLD
    & 8.929 & 2.823\% & 4s
    & 14.835 & 3.214\% & 14s
    \\
    & CaDA
    & 8.165 & 1.477\% & 3s
    & 11.992 & 0.960\% & 12s
    & & CaDA
    & 8.901 & 2.482\% & 4s
    & 14.747 & 2.615\% & 17s
    \\
    & CCL
    & 8.184 & 1.715\% & 6s
    & 12.039 & 1.355\% & 19s
    & & CCL
    & \textbf{8.870} & \textbf{2.138\%} & 7s
    & 14.723 & 2.436\% & 23s
    \\
    \rowcolor{latrow}
    & LaT-ReLD
    & \textbf{8.152} & \textbf{1.313\%} & 3s
    & \textbf{11.981} & \textbf{0.867\%} & 10s
    & & LaT-ReLD
    & 8.900 & 2.480\% & 4s
    & 14.765 & 2.737\% & 14s
    \\
    \rowcolor{latrow}
    & LaT-CaDA
    & 8.154 & 1.330\% & 3s
    & 11.984 & 0.892\% & 12s
    & & LaT-CaDA
    & 8.879 & 2.232\% & 4s
    & \textbf{14.722} & \textbf{2.431\%} & 17s
    \\
    \bottomrule
  \end{tabular}%
  }
  \end{small}
\end{table*}

\subsection{Ablation Study}
\label{sec:component-ablation}

\begin{table*}[!t]
  \centering
  \caption{Performance of different methods for zero-shot generalization
  on 1K unseen VRP instances. The best results among the multi-task
  neural solvers are highlighted in \textbf{bold}.}
  \label{tab:2}

  \begin{small}
  \setlength{\tabcolsep}{2.8pt}
  \renewcommand{\arraystretch}{1.05}

  \resizebox{\textwidth}{!}{%
  \begin{tabular}{
    @{}llcccccc
    @{\hspace{8pt}}
    llcccccc@{}
  }
    \toprule
    \multicolumn{2}{c}{\multirow{2}{*}{Method}}
    & \multicolumn{3}{c}{$N=50$}
    & \multicolumn{3}{c}{$N=100$}
    & \multicolumn{2}{c}{\multirow{2}{*}{Method}}
    & \multicolumn{3}{c}{$N=50$}
    & \multicolumn{3}{c}{$N=100$}
    \\
    \cmidrule(lr){3-5}
    \cmidrule(lr){6-8}
    \cmidrule(lr){11-13}
    \cmidrule(lr){14-16}
    & & Obj. & Gap & Time
    & Obj. & Gap & Time
    & && Obj. & Gap & Time
    & Obj. & Gap & Time
    \\
    \midrule

    \multirow{12}{*}{\rotatebox{90}{OVRPB}}
    & OR-Tools
    & 5.764 & 0.332\% & 10.4m
    & 8.522 & 1.852\% & 20.8m
    & \multirow{12}{*}{\rotatebox{90}{OVRPL}}
    & OR-Tools
    & 6.522 & 0.480\% & 10.4m
    & 9.966 & 1.783\% & 20.8m
    \\
    & OR-Tools ($\times 10$)
    & 5.745 & 0.000\% & 1.7h
    & 8.365 & 0.000\% & 3.5h
    & & OR-Tools ($\times 10$)
    & 6.490 & 0.000\% & 1.7h
    & 9.790 & 0.000\% & 3.5h
    \\
    & POMO-MTL
    & 6.116 & 6.430\% & 2s
    & 8.979 & 7.335\% & 8s
    & & POMO-MTL
    & 6.668 & 2.734\% & 2s
    & 10.126 & 3.441\% & 10s
    \\
    & MVMoE
    & 6.092 & 5.999\% & 3s
    & 8.959 & 7.088\% & 11s
    & & MVMoE
    & 6.650 & 2.454\% & 3s
    & 10.097 & 3.148\% & 13s
    \\
    & MVMoE-L
    & 6.122 & 6.522\% & 3s
    & 8.972 & 7.243\% & 10s
    & & MVMoE-L
    & 6.659 & 2.597\% & 3s
    & 10.106 & 3.244\% & 12s
    \\
    & RF-POMO
    & 6.077 & 5.746\% & 2s
    & 8.912 & 6.529\% & 8s
    & & RF-POMO
    & 6.647 & 2.413\% & 2s
    & 10.088 & 3.044\% & 10s
    \\
    & SPSM
    & 6.061 & 5.456\% & 3s
    & 8.863 & 5.945\% & 12s
    & & SPSM
    & 6.639 & 2.291\% & 3s
    & 10.055 & 2.724\% & 14s
    \\
    & ReLD
    & 6.044 & 5.166\% & 3s
    & 8.825 & 5.495\% & 10s
    & & ReLD
    & 6.638 & 2.264\% & 3s
    & 10.026 & 2.427\% & 12s
    \\
    & CaDA
    & 6.064 & 5.509\% & 3s
    & 8.968 & 7.183\% & 12s
    & & CaDA
    & 6.616 & 1.924\% & 3s
    & 10.000 & 2.145\% & 14s
    \\
    & CCL
    & 6.058 & 5.413\% & 6s
    & 8.864 & 5.962\% & 18s
    & & CCL
    & 6.612 & 1.880\% & 6s
    & 10.044 & 2.594\% & 20s
    \\
    \rowcolor{latrow}
    & LaT-ReLD
    & \textbf{6.033} & \textbf{4.982\%} & 3s
    & \textbf{8.800} & \textbf{5.196\%} & 10s
    & & LaT-ReLD
    & 6.622 & 2.021\% & 3s
    & 10.005 & 2.196\% & 12s
    \\
    \rowcolor{latrow}
    & LaT-CaDA
    & 6.040 & 5.104\% & 3s
    & 8.848 & 5.753\% & 12s
    & & LaT-CaDA
    & \textbf{6.606} & \textbf{1.773\%} & 3s
    & \textbf{9.997} & \textbf{2.114\%} & 14s
    \\
    \midrule

    \multirow{12}{*}{\rotatebox{90}{VRPBL}}
    & OR-Tools
    & 8.131 & 1.254\% & 10.4m
    & 12.095 & 2.586\% & 20.8m
    & \multirow{12}{*}{\rotatebox{90}{VRPBTW}}
    & OR-Tools
    & 15.053 & 1.857\% & 10.4m
    & 26.217 & 2.828\% & 20.8m
    \\
    & OR-Tools ($\times 10$)
    & 8.029 & 0.000\% & 1.7h
    & 11.790 & 0.000\% & 3.5h
    & & OR-Tools ($\times 10$)
    & 14.771 & 0.000\% & 1.7h
    & 25.496 & 0.000\% & 3.5h
    \\
    & POMO-MTL
    & 8.188 & 1.971\% & 2s
    & 11.998 & 1.764\% & 9s
    & & POMO-MTL
    & 16.055 & 8.841\% & 3s
    & 27.319 & 7.413\% & 11s
    \\
    & MVMoE
    & 8.172 & 1.776\% & 3s
    & 11.945 & 1.315\% & 11s
    & & MVMoE
    & 16.022 & 8.600\% & 3s
    & 27.236 & 7.078\% & 14s
    \\
    & MVMoE-L
    & 8.180 & 1.872\% & 3s
    & 11.960 & 1.442\% & 10s
    & & MVMoE-L
    & 16.041 & 8.745\% & 3s
    & 27.265 & 7.190\% & 13s
    \\
    & RF-POMO
    & 8.180 & 1.869\% & 2s
    & 11.951 & 1.366\% & 9s
    & & RF-POMO
    & 16.024 & 8.636\% & 3s
    & 27.210 & 6.982\% & 11s
    \\
    & SPSM
    & 8.172 & 1.781\% & 3s
    & 11.932 & 1.204\% & 12s
    & & SPSM
    & 16.000 & 8.464\% & 3s
    & 27.199 & 6.936\% & 15s
    \\
    & ReLD
    & 8.168 & 1.723\% & 3s
    & 11.918 & 1.086\% & 10s
    & & ReLD
    & 16.012 & 8.549\% & 2s
    & 27.137 & 6.691\% & 13s
    \\
    & CaDA
    & 8.178 & 1.855\% & 3s
    & 11.919 & 1.094\% & 11s
    & & CaDA
    & 16.051 & 8.803\% & 3s
    & 27.188 & 6.899\% & 15s
    \\
    & CCL
    & 8.186 & 1.958\% & 6s
    & 11.955 & 1.440\% & 17s
    & & CCL
    &15.971 & 8.261\%& 7s
    & 27.135 & 6.679\% & 22s
    \\
    \rowcolor{latrow}
    & LaT-ReLD
    & 8.167 & 1.717\% & 3s
    & 11.910 & 1.018\% & 10s
    & & LaT-ReLD
    & 16.030 & 8.661\% & 2s
    & 27.198 & 6.955\% & 13s
    \\
    \rowcolor{latrow}
    & LaT-CaDA
    & \textbf{8.161} & \textbf{1.643\%} & 3s
    & \textbf{11.909 }& \textbf{1.009\%} & 12s
    & & LaT-CaDA
    &  \textbf{15.970 }&\textbf{ 8.225\% } & 3s
    & \textbf{27.098} & \textbf{6.523\%} & 15s
    \\
    \midrule

    \multirow{12}{*}{\rotatebox{90}{VRPLTW}}
    & OR-Tools
    & 14.815 & 1.432\% & 10.4m
    & 25.823 & 2.534\% & 20.8m
    & \multirow{12}{*}{\rotatebox{90}{OVRPBL}}
    & OR-Tools
    & 5.771 & 0.549\% & 10.4m
    & 8.555 & 2.459\% & 20.8m
    \\
    & OR-Tools ($\times 10$)
    & 14.598 & 0.000\% & 1.7h
    & 25.195 & 0.000\% & 3.5h
    & & OR-Tools ($\times 10$)
    & 5.739 & 0.000\% & 1.7h
    & 8.348 & 0.000\% & 3.5h
    \\
    & POMO-MTL
    & 14.961 & 2.586\% & 3s
    & 25.619 & 1.920\% & 13s
    & & POMO-MTL
    & 6.104 & 6.306\% & 2s
    & 8.961 & 7.343\% & 9s
    \\
    & MVMoE
    & 14.937 & 2.421\% & 4s
    & 25.514 & 1.471\% & 17s
    & & MVMoE
    & 6.076 & 5.843\% & 3s
    & 8.942 & 7.115\% & 12s
    \\
    & MVMoE-L
    & 14.953 & 2.535\% & 4s
    & 25.529 & 1.545\% & 16s
    & & MVMoE-L
    & 6.104 & 6.310\% & 3s
    & 8.957 & 7.300\% & 11s
    \\
    & RF-POMO
    & 14.933 & 2.392\% & 3s
    & 25.530 & 1.552\% & 13s
    & & RF-POMO
    & 6.062 & 5.586\% & 2s
    & 8.894 & 6.535\% & 9s
    \\
    & SPSM
    & 14.924 & 2.334\% & 4s
    & 25.474 & 1.325\% & 18s
    & & SPSM
    & 6.051 & 5.401\% & 3s
    & 8.848 & 5.984\% & 14s
    \\
    & ReLD
    & 14.921 & 2.294\% & 4s
    & 25.449 & 1.218\% & 16s
    & & ReLD
    & 6.034 & 5.103\% & 3s
    & 8.806 & 5.492\% & 11s
    \\
    & CaDA
    & 14.895 & 2.127\% & 4s
    & 25.399 & 1.026\% & 18s
    & & CaDA
    & 6.047 & 5.338\% & 3s
    & 8.961 & 7.326\% & 14s
    \\
    & CCL
    & \textbf{14.856} &\textbf{ 1.851\% }& 7s
    & 25.366 & 0.882\% & 25s
    & & CCL
    & 6.041 & 5.234\% & 6s
    & 8.851 & 6.033\% & 21s
    \\
    \rowcolor{latrow}
    & LaT-ReLD
    & 14.893 & 2.107\% & 4s
    & 25.392 & 0.996\% & 16s
    & & LaT-ReLD
    & \textbf{6.022} & \textbf{4.890\%} & 3s
    & \textbf{8.789} & \textbf{5.275\%} & 11s
    \\
    \rowcolor{latrow}
    & LaT-CaDA
    & 14.864 &1.909\% & 4s
    & \textbf{25.357} & \textbf{0.838\%} & 18s
    & & LaT-CaDA
    & 6.025 & 4.946\% & 3s
    & 8.834 & 5.813\% & 14s
    \\
    \midrule

    \multirow{12}{*}{\rotatebox{90}{OVRPBTW}}
    & OR-Tools
    & 8.758 & 0.927\% & 10.4m
    & 14.713 & 2.268\% & 20.8m
    & \multirow{12}{*}{\rotatebox{90}{OVRPLTW}}
    & OR-Tools
    & 8.728 & 0.656\% & 10.4m
    & 14.535 & 1.779\% & 20.8m
    \\
    & OR-Tools ($\times 10$)
    & 8.675 & 0.000\% & 1.7h
    & 14.384 & 0.000\% & 3.5h
    & & OR-Tools ($\times 10$)
    & 8.669 & 0.000\% & 1.7h
    & 14.279 & 0.000\% & 3.5h
    \\
    & POMO-MTL
    & 9.514 & 9.628\% & 3s
    & 15.879 & 10.453\% & 10s
    & & POMO-MTL
    & 8.987 & 3.633\% & 3s
    & 14.896 & 4.374\% & 12s
    \\
    & MVMoE
    & 9.486 & 9.308\% & 4s
    & 15.808 & 9.948\% & 13s
    & & MVMoE
    & 8.966 & 3.396\% & 4s
    & 14.828 & 3.903\% & 15s
    \\
    & MVMoE-L
    & 9.515 & 9.630\% & 3s
    & 15.841 & 10.188\% & 12s
    & & MVMoE-L
    & 8.974 & 3.488\% & 4s
    & 14.839 & 3.971\% & 14s
    \\
    & RF-POMO
    & 9.488 & 9.327\% & 3s
    & 15.787 & 9.808\% & 10s
    & & RF-POMO
    & 8.950 & 3.213\% & 3s
    & 14.820 & 3.840\% & 12s
    \\
    & SPSM
    & 9.475 & 9.180\% & 4s
    & 15.754 & 9.578\% & 15s
    & & SPSM
    & 8.941 & 3.119\% & 4s
    & 14.772 & 3.505\% & 17s
    \\
    & ReLD
    & 9.467 & 9.087\% & 3s
    & 15.721 & 9.349\% & 12s
    & & ReLD
    & 8.932 & 3.013\% & 4s
    & 14.742 & 3.302\% & 14s
    \\
    & CaDA
    & 9.480 & 9.229\% & 4s
    & 15.710 & 9.285\% & 15s
    & & CaDA
    & 8.897 & 2.601\% & 4s
    & 14.633 & 2.544\% & 17s
    \\
    & CCL
    & \textbf{9.422} & \textbf{8.581\%} & 7s
    & \textbf{15.666 }& \textbf{8.966\% }& 22s
    & & CCL
    &\textbf{8.864} &\textbf{ 2.228\% }& 7s
    & 14.627 & 2.498\% & 24s
    \\
    \rowcolor{latrow}
    & LaT-ReLD
    & 9.451 & 8.904\% & 3s
    & 15.739 & 9.482\% & 12s
    & & LaT-ReLD
    & 8.899 & 2.622\% & 4s
    & 14.677 & 2.848\% & 14s
    \\
    \rowcolor{latrow}
    & LaT-CaDA
    & 9.432 & 8.684\% & 4s
    & 15.671 & 9.008\% & 15s
    & & LaT-CaDA
    & 8.877 & 2.380\% & 4s
    & \textbf{14.615} & \textbf{2.413\%} & 17s
    \\
    \midrule

    \multirow{12}{*}{\rotatebox{90}{VRPBLTW}}
    & OR-Tools
    & 14.890 & 1.402\% & 10.4m
    & 25.979 & 2.518\% & 20.8m
    & \multirow{12}{*}{\rotatebox{90}{OVRPBLTW}}
    & OR-Tools
    & 8.729 & 0.624\% & 10.4m
    & 14.496 & 1.724\% & 20.8m
    \\
    & OR-Tools ($\times 10$)
    & 14.677 & 0.000\% & 1.7h
    & 25.342 & 0.000\% & 3.5h
    & & OR-Tools ($\times 10$)
    & 8.673 & 0.000\% & 1.7h
    & 14.250 & 0.000\% & 3.5h
    \\
    & POMO-MTL
    & 15.980 & 9.035\% & 3s
    & 27.247 & 7.746\% & 12s
    & & POMO-MTL
    & 9.532 & 9.851\% & 4s
    & 15.738 & 10.498\% & 15s
    \\
    & MVMoE
    & 15.945 & 8.775\% & 4s
    & 27.142 & 7.332\% & 14s
    & & MVMoE
    & 9.503 & 9.516\% & 3s
    & 15.671 & 9.972\% & 12s
    \\
    & MVMoE-L
    & 15.963 & 8.915\% & 4s
    & 27.177 & 7.473\% & 14s
    & & MVMoE-L
    & 9.518 & 9.682\% & 4s
    & 15.706 & 10.263\% & 13s
    \\
    & RF-POMO
    & 15.965 & 8.910\% & 3s
    & 27.132 & 7.291\% & 12s
    & & RF-POMO
    & 9.497 & 9.458\% & 3s
    & 15.661 & 9.948\% & 11s
    \\
    & SPSM
    & 15.939 & 8.744\% & 4s
    & 27.086 & 7.102\% & 16s
    & & SPSM
    & 9.480 & 9.260\% & 4s
    & 15.621 & 9.672\% & 15s
    \\
    & ReLD
    & 15.917 & 8.589\% & 4s
    & 27.062 & 7.014\% & 14s
    & & ReLD
    & 9.480 & 9.254\% & 4s
    & 15.587 & 9.433\% & 13s
    \\
    & CaDA
    & 15.968 & 8.917\% & 4s
    & 27.102 & 7.196\% & 16s
    & & CaDA
    & 9.489 & 9.357\% & 4s
    & 15.586 & 9.440\% & 15s
    \\
    & CCL
    & 15.910 & 8.507\% & 7s
    & 27.064 & 7.004\% & 23s
    & & CCL
    &\textbf{ 9.433} & \textbf{8.726\%} & 8s
    & \textbf{15.522} & \textbf{8.969\% }& 22s
    \\
    \rowcolor{latrow}
    & LaT-ReLD
    & 15.958 & 8.836\% & 4s
    & 27.157 & 7.380\% & 14s
    & & LaT-ReLD
    & 9.473 & 9.158\% & 4s
    & 15.603 & 9.549\% & 13s
    \\
    \rowcolor{latrow}
    & LaT-CaDA
    & \textbf{15.891} & \textbf{8.380\%} & 4s
    & \textbf{27.044} & \textbf{6.950\%} & 16s
    & & LaT-CaDA
    & 9.445 & 8.859\% & 4s
    & 15.528 & 9.013\% & 15s
    \\
    \bottomrule
  \end{tabular}%
  }
  \end{small}
\end{table*}

To examine whether the improvement arises from the LLM-generated guidance vector
rather than merely from the additional parameters introduced by the encoder
side branches, we conduct a component ablation study using LaT with the
POMO-MTL backbone on 50-node multi-task VRPs. The two vector ablations retain
the same encoder side branches and ten-dimensional conditioning input
\(\mathbf{c}_{k,r}=[\mathbf{z}_k,\boldsymbol{\gamma}_r]\).
In the fixed-guidance variant, the LLM-generated guidance vector is replaced with
\(\mathbf{1}_5\) throughout training, while the task constraint vector remains
unchanged. In the no-constraint variant, the task constraint vector is replaced
with \(\mathbf{0}_5\), while the LLM-generated guidance vector is retained.
Multiplicative fusion further replaces concatenation with the element-wise
product of the two vectors. Thus, the two vector ablations have the same input
dimension, side-branch architecture, and number of trainable parameters as
LaT.

\begin{figure*}[!htbp]
    \centering
    \includegraphics[width=0.98\textwidth]
    {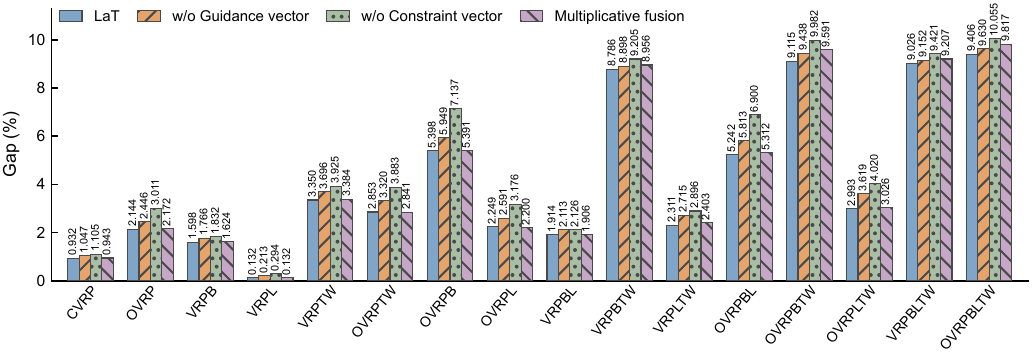}
    \caption{Component ablation of LaT with the POMO-MTL backbone on
    50-node multi-task VRPs. Lower values indicate better performance.}
    \label{fig:component-ablation}
\end{figure*}

As shown in Fig.~\ref{fig:component-ablation}, LaT achieves the lowest average
gap and outperforms both vector ablations across all \(16\) VRP variants.
Replacing the LLM-generated guidance with a fixed vector consistently degrades
solution quality, although the model architecture and parameter count remain
unchanged. This controlled comparison confirms that the improvement is
attributable to the task-adaptive guidance generated by the LLM, rather than
solely to the increased capacity of the side branches. Removing the task
constraint vector also worsens performance, demonstrating that explicit task
information provides a complementary contribution. Moreover, multiplicative
fusion yields a higher average gap than concatenation, indicating that
concatenation better preserves and integrates the two sources of information.
Although LaT increases the trainable parameter count only moderately from
approximately \(1.25\)M to \(1.46\)M, these results demonstrate that its gains
primarily arise from the LLM-generated guidance vector, the task constraint vector,
and their effective fusion, rather than from the parameter increase alone.
Further details on the model complexity and parameter count are provided in
Appendix~\ref{app:complexity-analysis}. Further evidence is provided by the MLP comparison in
Appendix~\ref{app:mlp-guidance-comparison}, where an MLP guidance generator
with additional trainable parameters yields a higher average gap
 than LaT, further confirming that the improvement
comes from the LLM-generated guidance vector rather than increased model capacity.



\section{Conclusion}

The goal of this paper is to address the lack of stage-wise task performance
feedback during the joint training of multi-task neural solvers. To this end,
we propose \textbf{LaT}, a plug-and-play training paradigm that uses a
pretrained LLM as an external trainer. LaT periodically analyzes cross-task
validation metrics to generate a guidance vector.
The vector is combined with the task constraint vector and provided to
lightweight side branches after each encoder layer, providing the neural solver
with additional stage-wise training information during subsequent policy
optimization. The LLM remains frozen, is queried only at predefined intervals,
and is removed after training, while the final guidance vector is fixed for
inference.  Experiments on 16 VRP variants
show that LaT improves the overall solution quality of multiple
state-of-the-art multi-task neural solvers on both trained and unseen variants.
Results on selected CVRPLIB Set-X and Set-Solomon instances further support its
generalization to benchmark instances.

Although LaT achieves favorable performance, several aspects remain to be
explored. First, LaT-CaDA shows more consistent improvements than LaT-ReLD on
unseen VRPs. This suggests that the effectiveness of LaT can depend on the
selected backbone. Developing guidance mechanisms that work more consistently
with different neural solvers is therefore a worthwhile direction. Second, the
current experiments cover 16 VRP variants formed from five elementary
constraints and two problem sizes. Extending LaT to additional VRP constraints
and more complex problem settings could further evaluate its generality and
applicability.



\bibliography{main}
\bibliographystyle{iclr2027_conference}

\newpage
\appendix
\section{Related Work}

\paragraph{Neural solvers for single-task VRPs.}
Early neural solvers usually formulate VRPs as sequential decision-making problems and construct a route autoregressively. Pointer Networks first demonstrated the potential of attention-based sequence models for combinatorial optimization \citep{vinyals2015pointer}, and subsequent reinforcement learning methods extended this paradigm to routing problems without requiring optimal labels \citep{bello2017neural,nazari2018reinforcement,joshi2019efficient}. The Attention Model further established a Transformer-style encoder--decoder framework for routing, where node embeddings are produced by the encoder and the decoder selects the next node step by step \citep{kool2018attention}. POMO improves this framework by exploiting multiple equivalent optima and multi-start rollouts, becoming a strong baseline for TSP and CVRP \citep{kwon2020pomo}. Other works enhance single-task neural solvers through iterative improvement, variant-specific modeling, or better attention/generalization mechanisms, such as learning-based iterative search for VRPs, joint attention for VRPTW, and distance-aware attention reshaping for improving solver generalization \citep{lu2020learning,falkner2020learning,hottung2020neural,xin2021neurolkh,ma2021learning,luo2023neural,invit,hua2025camp,wang2025dar}. Although these methods achieve strong performance on individual variants, they are usually designed for a fixed problem setting and must be retrained or redesigned when the constraint set changes.

\paragraph{Neural solvers for multi-task VRPs.}
To avoid training a separate model for each VRP variant, recent studies have
developed unified neural solvers that share knowledge across constraints and
tasks \citep{wang2024asp,drakulic2024goal,zong2025unico}. Cross-problem learning
and soft parameter sharing show that knowledge learned from one routing variant
can be transferred to related variants
\citep{lin2024cross,huang2025rethinking,wang2026soft}. POMO-MTL represents each
task as a composition of elementary constraints and supports zero-shot
generalization to unseen constraint combinations \citep{mtl}. MVMoE increases
model capacity through expert specialization, RouteFinder trains
Transformer-based models in a unified VRP environment, CaDA improves constraint
awareness through constraint prompts and dual attention, and SHIELD enhances
multi-task and multi-distribution generalization through sparse and hierarchical
computation \citep{zhou2024mvmoe,berto2025routefinder,li2024cada,shield}. Recent
studies further explore task specialization, context modeling, and policy
optimization. MoSES reuses specialized basis policies under a
State-Decomposable MDP \citep{moses}. Mixed-curvature pre-training captures
heterogeneous geometric structures across tasks \citep{liu2025mixed}, while
MTL-KD distills multiple single-task teachers into a multi-task solver
\citep{zheng2025mtlkd}. CCL updates the decoding context and node
representations according to evolving constraint relevance
\citep{gui2026ccl}, and PoMtVRS combines preference-gated decoding with
within-instance preference learning \citep{meng2026pomt}. Recent studies have incorporated LLMs into solvers for VRPs in different
ways. Some methods use LLM-derived semantics to enrich or align spatial and
graph representations \citep{malik2026llmaide,feng2026aligning}. DRoC
decomposes complex constraints and retrieves constraint-specific knowledge to
help LLMs exploit existing solvers \citep{jiang2025droc}. Other methods use
LLMs to generate or evolve attention-shaping heuristics for solver fine-tuning
and variant adaptation \citep{tran2025large,chi2026generalized}.
 Despite this progress, existing methods do not explicitly use stage-wise
cross-task validation performance during joint training. In contrast,
\textbf{LLM-as-Trainer (LaT)} uses a pretrained LLM as an external trainer.
At predefined intervals, the LLM analyzes cross-task validation metrics to generate a guidance vector. After training, the
final guidance vector is fixed and the LLM is removed, requiring no LLM API
calls at inference.

\section{Multi-Task VRP Details}
\label{app:mtvrp-details}

\subsection{VRP Variants}
\label{app:vrp-variants}
We consider a family of \(16\) VRP variants constructed from the five
elementary constraints in
\(\mathcal{M}=\{\mathrm{C},\mathrm{O},\mathrm{B},\mathrm{L},\mathrm{TW}\}\),
namely capacity, open route, backhaul, route-length limit, and time window.
CVRP is the base problem, and the remaining variants are obtained by activating
different subsets of \(\{\mathrm{O},\mathrm{B},\mathrm{L},\mathrm{TW}\}\) in
addition to the capacity constraint. Consistent with the main text, each task
\(k\) is represented by the constraint vector \(\mathbf{z}_k\) defined in
Eq.~\ref{eq:zvec}, and an active entry enforces the corresponding constraint
through action masking. Table~\ref{tab:vrp-variants} lists the \(16\) tasks.

\begin{table}[t]
    \centering
    \caption{The 16 VRP variants and their active constraints. Capacity
    is included in every task.}
    \label{tab:vrp-variants}
    \small
    \setlength{\tabcolsep}{5pt}
    \begin{tabular}{lccccc}
        \toprule
        Variant & C & O & B & L & TW \\
        \midrule
        CVRP       & $\checkmark$ &              &              &              &              \\
        OVRP       & $\checkmark$ & $\checkmark$ &              &              &              \\
        VRPB       & $\checkmark$ &              & $\checkmark$ &              &              \\
        VRPL       & $\checkmark$ &              &              & $\checkmark$ &              \\
        VRPTW      & $\checkmark$ &              &              &              & $\checkmark$ \\
        OVRPTW     & $\checkmark$ & $\checkmark$ &              &              & $\checkmark$ \\
        OVRPB      & $\checkmark$ & $\checkmark$ & $\checkmark$ &              &              \\
        OVRPL      & $\checkmark$ & $\checkmark$ &              & $\checkmark$ &              \\
        VRPBL      & $\checkmark$ &              & $\checkmark$ & $\checkmark$ &              \\
        VRPBTW     & $\checkmark$ &              & $\checkmark$ &              & $\checkmark$ \\
        VRPLTW     & $\checkmark$ &              &              & $\checkmark$ & $\checkmark$ \\
        OVRPBL     & $\checkmark$ & $\checkmark$ & $\checkmark$ & $\checkmark$ &              \\
        OVRPBTW    & $\checkmark$ & $\checkmark$ & $\checkmark$ &              & $\checkmark$ \\
        OVRPLTW    & $\checkmark$ & $\checkmark$ &              & $\checkmark$ & $\checkmark$ \\
        VRPBLTW    & $\checkmark$ &              & $\checkmark$ & $\checkmark$ & $\checkmark$ \\
        OVRPBLTW   & $\checkmark$ & $\checkmark$ & $\checkmark$ & $\checkmark$ & $\checkmark$ \\
        \bottomrule
    \end{tabular}
\end{table}

\subsection{VRP Instance Generation}
\label{app:vrp-generation}
A VRP instance is defined on a complete graph \(G=(\mathcal{V},\mathcal{E})\) with
\(\mathcal{V}=\{v_0,v_1,\ldots,v_N\}\), one depot \(v_0\) and \(N\) customers. Node
\(v_i\) has coordinates \(\mathbf{x}_i=(x_{i,1},x_{i,2})\), and the travel cost is the
Euclidean distance \(d_{ij}=\lVert\mathbf{x}_i-\mathbf{x}_j\rVert_2\). A solution \(\tau\) visits every customer exactly once while satisfying the
constraints of task \(k\), and the objective is to minimize the total distance
over the feasible set \(\mathcal{F}_k\):
\begin{equation}
    \tau_k^{\star}=\arg\min_{\tau\in\mathcal{F}_k} c(\tau),
\end{equation}
where the final customer-to-depot edge is omitted for open-route tasks.

\paragraph{Locations and travel times.}
Coordinates are sampled independently from the unit square,
\(\mathbf{x}_i\sim\mathcal{U}([0,1]^2)\) for \(i=0,\ldots,N\). The vehicle speed
is one, so the travel time equals the Euclidean distance.

\paragraph{Capacity constraint (C).}
Each customer receives an integer demand
\(\widetilde q_i\sim\mathcal{U}\{1,\ldots,9\}\), and the capacity depends on the
problem size:
\begin{equation}
Q=
\begin{cases}
30, & N=20,\\
40, & N=50,\\
50, & N=100,\\
70, & N=200.
\end{cases}
\end{equation}
The normalized demand is \(q_i=\widetilde q_i/Q\), the normalized capacity
equals one, and a customer is feasible only if its demand does not exceed the
remaining load.

\paragraph{Open-route constraint (O).}
For closed-route tasks every route starts and ends at the depot, whereas for
open-route tasks a route may terminate at its last visited customer, and the
final customer-to-depot edge is excluded from both the cost and the
route-length feasibility check.

\paragraph{Backhaul constraint (B).}
For tasks with backhauls, \(\lfloor 0.2N\rfloor\) customers are randomly
selected as backhaul customers and their normalized demands take negative
signs:
\begin{equation}
q_i=
\begin{cases}
 \widetilde q_i/Q,  & i\in\mathcal{V}_{\mathrm{LH}},\\
-\widetilde q_i/Q,  & i\in\mathcal{V}_{\mathrm{BH}},
\end{cases}
\end{equation}
where \(\mathcal{V}_{\mathrm{LH}}\) and \(\mathcal{V}_{\mathrm{BH}}\) denote the
linehaul and the backhaul customers, and the normalized load must stay within
\([0,1]\) throughout the construction.

\paragraph{Route-length constraint (L).}
For tasks with a route-length limit, each route is bounded by \(R=3.0\). For a
partial closed route ending at \(v_i\) with accumulated route length \(D_t\),
selecting \(v_j\) is feasible only if \(D_t+d_{ij}+d_{j0}\le R\), and the return
term \(d_{j0}\) is removed for open-route tasks. The quantity \(D_t\) is exactly
the third entry of the dynamic state \(\mathbf{d}_t\) used by the decoder in
Eq.~\ref{eq:context}.

\paragraph{Time-window constraint (TW).}
The depot window is \([0,T]\) with \(T=3\), and every customer has a service
duration \(\delta_i=0.2\). Using the distance \(d_{0i}\) from the depot to
customer \(v_i\), the sampling bounds of the window center are:
\begin{equation}
    \underline{\mu}_i=d_{0i},\qquad \overline{\mu}_i=T-d_{0i}-\delta_i.
\end{equation}
A window center and a half-width are sampled as:
\(\mu_i\sim\mathcal{U}(\underline{\mu}_i,\overline{\mu}_i)\) and
\(w_i\sim\mathcal{U}(\delta_i/2,\,T/3)\), and the time window is
\begin{equation}
    e_i=\max(0,\mu_i-w_i),\qquad l_i=\min(T,\mu_i+w_i).
\end{equation}
If the service completes at time \(\vartheta\) at node \(v_i\), the service start time
at a candidate node \(v_j\) is \(\max(\vartheta+d_{ij},e_j)\), and the action is
feasible only if this start time does not exceed \(l_j\). Closed-route tasks
additionally require returning to the depot before \(T\).

\paragraph{Composition of variants.}
All variants share the same coordinate and demand generation, and a task is
instantiated by activating the entries of \(\mathbf{z}_k\) and enforcing the
associated constraints through action masking. Inactive time-window attributes
are set to zero, and backhaul demands stay positive when the backhaul constraint is inactive.

\section{Prompts for Guidance Generation}
\label{app:prompts}

This appendix presents the prompts used by the pretrained LLM to generate the
guidance vector. At each guidance update, the user message provides the
constraint compositions and recent validation results of all tasks. The system
message and response format remain unchanged, while the structured training
state is updated with the latest validation records. The same prompt format is
used in all multi-task VRP settings, and the LLM temperature is set to zero.
Section~\ref{app:prompt-format} presents the prompt format, and
Section~\ref{app:control-dynamics} reports the guidance vectors generated for
the 50-node POMO-MTL model.

\subsection{Prompt Format}
\label{app:prompt-format}

The system message specifies the role of the LLM, the order of the five
guidance dimensions, and the required output format.

\begin{promptbox}
You are an LLM trainer for a multi-task VRP model. Analyze validation score/gap and return strict JSON only. The guidance vector order is [C, O, B, L, TW], meaning capacity, open route, backhaul, route length, and time window. Values must be conservative floats.
\end{promptbox}
\promptcaption{System message defining the role of the LLM and the guidance-vector format.}{prompt:system}

The user message provides the structured training state
\(\mathcal{I}_r\). The fields \texttt{base\_guidance} and
\texttt{current\_guidance} denote the initial guidance vector
\(\boldsymbol{\gamma}_0\) and the vector used during the preceding training
interval, respectively. Each task record contains its active constraints,
latest validation objective, relative reference gap, gap change, and up to
\(J\) recent validation records. The values in angle brackets are populated
at runtime.

\begin{promptbox}
Given this training state, output JSON with keys: guidance: array of 5 floats, reason: short string. Allowed guidance range is [0.5, 2.0]. For the first control round, previous_guidance equals initial_guidance and gap_delta is 0 for every task.

{
  "guidance_names": ["C", "O", "B", "L", "TW"],
  "initial_guidance": [
    <gamma0_C>,
    <gamma0_O>,
    <gamma0_B>,
    <gamma0_L>,
    <gamma0_TW>
  ],
  "previous_guidance": [
    <gamma_prev_C>,
    <gamma_prev_O>,
    <gamma_prev_B>,
    <gamma_prev_L>,
    <gamma_prev_TW>
  ],
  "tasks": [
    {
      "problem": "<VRP task name>",
      "score": <latest validation objective>,
      "gap": <latest relative reference gap>,
      "gap_delta": <latest gap minus previous gap, or 0 at the first control round>,
      "score_history": [
        <recent validation objectives>
      ],
      "gap_history": [
        <recent relative reference gaps>
      ]
    },
    ...
  ],
  "instruction": "Return conservative guidance values based on the current validation state of the five constraints."
}
\end{promptbox}
\promptcaption{User message containing cross-task validation results.}{prompt:user}

The LLM returns a five-dimensional guidance vector and a short explanation in the following JSON format:

\begin{promptbox}
{
 "guidance": [<gamma_r^C>, <gamma_r^O>, <gamma_r^B>, <gamma_r^L>, <gamma_r^TW>],
  "reason": "<short explanation>"
}
\end{promptbox}
\promptcaption{Required JSON output containing the guidance vector and a short explanation.}{prompt:response}

\subsection{Guidance update results on 50-node multi-task VRPs}
\label{app:control-dynamics}

\begin{figure*}[htpb]
  \centering
  \begin{minipage}[t]{0.66\textwidth}
    \vspace{0pt}
    \centering
    \includegraphics[width=\linewidth]
    {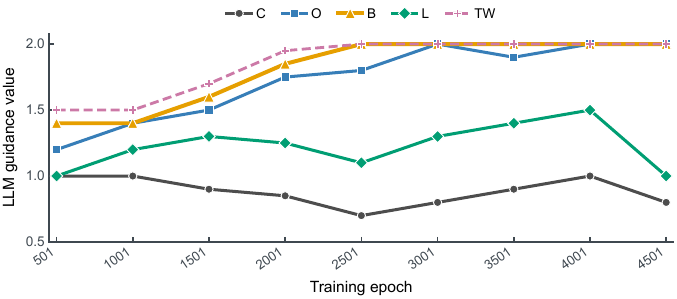}
    \par\smallskip
    {\small\textbf{(a)} guidance vector}
  \end{minipage}
  \hfill
  \begin{minipage}[t]{0.31\textwidth}
    \vspace{0pt}
    \centering
    \includegraphics[width=\linewidth]
    {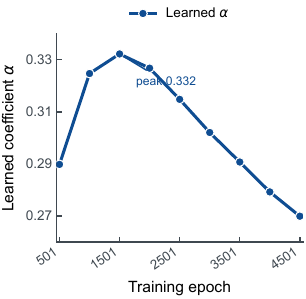}
    \par\smallskip
    {\small\textbf{(b)} Learned coefficient}
  \end{minipage}

\caption{Guidance dynamics of LaT-POMO-MTL on 50-node multi-task
VRPs. \textbf{(a)} Five-dimensional guidance vectors generated after
each completed control interval and applied from the epochs shown on
the horizontal axis. \textbf{(b)} Learned  coefficient
\(\alpha\) recorded at the same guidance updates.}
  \label{fig:n50-control-dynamics}
\end{figure*}

The recorded guidance updates are visualized below. They were obtained by
integrating LaT with the POMO-MTL backbone on 50-node multi-task
VRPs. The records in \texttt{control\_updates.jsonl} contain the
five-dimensional guidance vector returned by the LLM, its textual explanation,
and the current value of the learnable coefficient \(\alpha\).
Fig.~\ref{fig:n50-control-dynamics} shows the guidance values and
\(\alpha\) recorded at the nine LLM update rounds.

Fig.~\ref{fig:n50-control-dynamics} visualizes the nine guidance updates
recorded during training of the 50-node POMO-MTL model. The generated values
for the open-route (O), backhaul (B), and time-window (TW) dimensions generally
increased over training. B and TW reached the upper bound specified in the
prompt at epoch 2501 and remained at this value in the subsequent updates. O
first reached 2.0 at epoch 3001 and remained close to this value thereafter. In
contrast, the capacity (C) value remained between 0.7 and 1.0, while the
route-length (L) value varied between 1.0 and 1.5. These results show that the
five guidance dimensions followed different stage-wise trajectories rather
than changing uniformly.

The learned coefficient \(\alpha\) followed a different trajectory. It
increased from 0.290 at epoch 501 to a maximum of 0.332 at epoch 1501 and then
gradually decreased to 0.270 at epoch 4501. This coefficient is updated through
gradient-based training of the neural solver and is only recorded at the LLM
update epochs. Its variation should therefore not be interpreted as a decision
made by the LLM. The textual explanations in the update records are also
LLM-generated summaries of the supplied validation statistics rather than
independent evidence of causality. The effectiveness of the resulting guidance
is instead evaluated through relative reference gaps, comparisons with the
original backbone, and the ablation studies.

\subsection{Association between Guidance Values and Validation Gaps}
\label{app:guidance-gap-association}

We conduct this experiment on LaT-POMO-MTL trained on 50-node multi-task VRPs
to examine whether the guidance vector reflects validation performance
differences among the five constraint dimensions. The analysis uses the
guidance values and corresponding validation results from all nine control
rounds. For constraint
\(m\in\{\mathrm{C},\mathrm{O},\mathrm{B},\mathrm{L},\mathrm{TW}\}\), let
\(\mathcal{K}_{m}=\{k\mid z_k^{m}=1\}\) denote the validation tasks in which
constraint \(m\) is active. At control round \(r\), their mean relative reference gap is computed as:
\begin{equation}
    \bar{\eta}_{r}^{m}
    =
    \frac{1}{|\mathcal{K}_{m}|}
    \sum_{k\in\mathcal{K}_{m}}
    \eta_{r,k}.
    \label{eq:constraint-mean-gap}
\end{equation}

We then compare the five mean gaps with their corresponding guidance values
and calculate the Spearman correlation:
\begin{equation}
    \rho_r
    =
    \operatorname{Spearman}
    \left(
        \{\bar{\eta}_{r}^{m}\}_{m},
        \{\gamma_{r}^{m}\}_{m}
    \right).
    \label{eq:guidance-gap-correlation}
\end{equation}

\begin{figure*}[t]
    \centering

    \begin{minipage}[t]{0.31\textwidth}
        \centering
        \includegraphics[width=\linewidth]
        {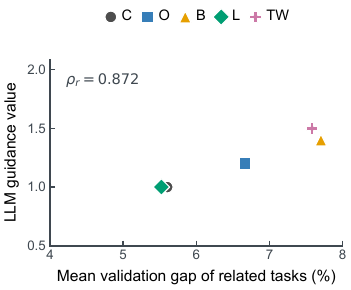}
        \par\smallskip
        \textbf{(a)} epoch 500
    \end{minipage}
    \hfill
    \begin{minipage}[t]{0.31\textwidth}
        \centering
        \includegraphics[width=\linewidth]
        {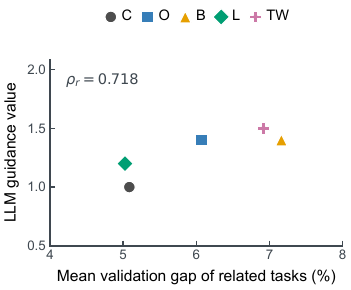}
        \par\smallskip
        \textbf{(b)} epoch 1000
    \end{minipage}
    \hfill
    \begin{minipage}[t]{0.31\textwidth}
        \centering
        \includegraphics[width=\linewidth]
        {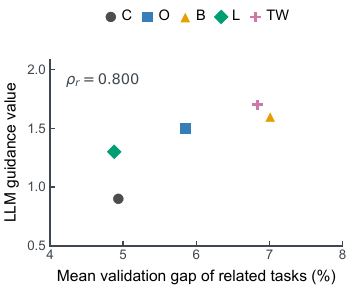}
        \par\smallskip
        \textbf{(c)} epoch 1500
    \end{minipage}

    \par\medskip

    \begin{minipage}[t]{0.31\textwidth}
        \centering
        \includegraphics[width=\linewidth]
        {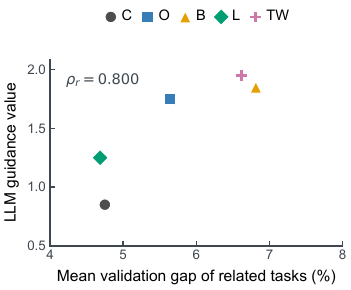}
        \par\smallskip
        \textbf{(d)} epoch 2000
    \end{minipage}
    \hfill
    \begin{minipage}[t]{0.31\textwidth}
        \centering
        \includegraphics[width=\linewidth]
        {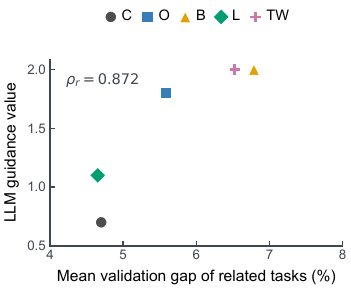}
        \par\smallskip
        \textbf{(e)} epoch 2500
    \end{minipage}
    \hfill
    \begin{minipage}[t]{0.31\textwidth}
        \centering
        \includegraphics[width=\linewidth]
        {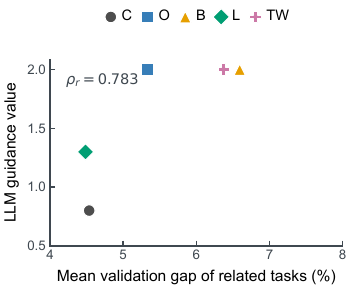}
        \par\smallskip
        \textbf{(f)} epoch 3000
    \end{minipage}

    \par\medskip

    \begin{minipage}[t]{0.31\textwidth}
        \centering
        \includegraphics[width=\linewidth]
        {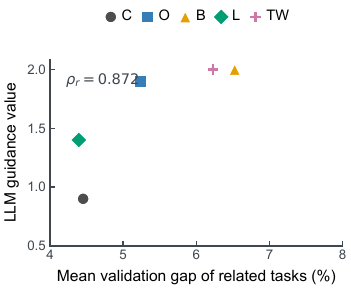}
        \par\smallskip
        \textbf{(g)} epoch 3500
    \end{minipage}
    \hfill
    \begin{minipage}[t]{0.31\textwidth}
        \centering
        \includegraphics[width=\linewidth]
        {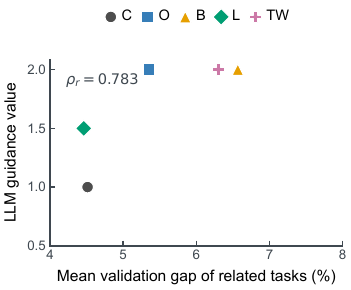}
        \par\smallskip
        \textbf{(h)} epoch 4000
    \end{minipage}
    \hfill
    \begin{minipage}[t]{0.31\textwidth}
        \centering
        \includegraphics[width=\linewidth]
        {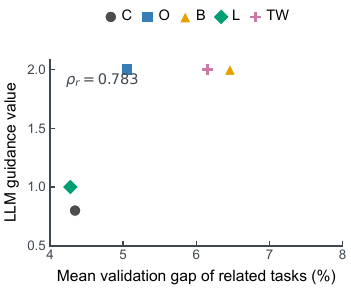}
        \par\smallskip
        \textbf{(i)} epoch 4500
    \end{minipage}

   \caption{Association between the LLM-generated guidance values and
constraint-related validation gaps on 50-node multi-task VRPs. For each
constraint, the mean gap is computed over the validation tasks in which that
constraint is active. Panels (a)--(i) show the nine control rounds, and
\(\rho_r\) denotes the Spearman correlation across the five constraint
dimensions at control round \(r\). At the same training stage, constraint
dimensions associated with larger mean validation gaps generally receive
larger guidance values.}
    \label{fig:guidance-gap-association}
\end{figure*}

As shown in Fig.~\ref{fig:guidance-gap-association}, the correlation is
positive in all nine control rounds, ranging from \(0.718\) to \(0.872\), with
an average of \(0.809\). These results show that, at the same training stage,
constraint dimensions associated with larger mean validation gaps generally
receive larger guidance values. This consistent association shows that the
guidance vector reflects current cross-task performance differences, supporting
the effectiveness of the proposed guidance mechanism.

\section{Backbone Details}
\label{app:backbone}

We describe the encoder--decoder backbone shared by the neural solvers
considered in this paper, following the Attention Model
\citep{kool2018attention} and POMO \citep{kwon2020pomo}. LaT retains the
backbone operations described below and adds a lightweight side branch to the
output \(\mathbf{h}_i^{\ell}\) of each encoder layer.

\begin{algorithm}[t]
\caption{LLM-as-Trainer (LaT)}
\label{alg:lat}
\begin{algorithmic}[1]

\Require solver \(\pi_\theta\) with encoder side branches and
coefficient \(\alpha\); pretrained LLM with prompt template
\(\mathcal{P}\); tasks \(\{\mathbf{z}_k\}_{k=1}^{16}\); validation
sets \(\{\mathcal{D}_k\}_{k=1}^{16}\); reference objectives
\(\{y_k^{\mathrm{ref}}\}_{k=1}^{16}\); guidance-update interval
\(T_{\mathrm{ctrl}}\); number of batches per epoch \(B\); total
number of epochs \(E\)

\Ensure trained solver \(\pi_\theta\) with fixed final guidance and
without LLM calls during inference

\State initialize \(W_{\mathrm{side}}^\ell\leftarrow\mathbf{0}\) for
\(\ell=1,\ldots,L_{\mathrm{enc}}\)
\State initialize \(\alpha\), guidance-round index \(r\leftarrow0\),
and initial guidance
\(\boldsymbol{\gamma}_0\leftarrow\mathbf{1}\)

\For{\(\mathrm{epoch}=1,2,\ldots,E\)}

    \For{\(b=1,2,\ldots,B\)}

        \State sample a training task \(k\) and generate a batch of
        instances under \(\mathbf{z}_k\)

        \State form
        \(\mathbf{u}_{r,k}\leftarrow
        [\mathbf{z}_k,\boldsymbol{\gamma}_r]\)

        \State embed the static node features:
        \[
        \mathbf{h}_i^0\leftarrow
        W_{\mathrm{in}}\mathbf{f}_i,
        \quad i=0,\ldots,N
        \]

        \State set
        \[
        \widetilde{\mathbf{h}}_i^0
        \leftarrow\mathbf{h}_i^0,
        \quad i=0,\ldots,N
        \]

        \For{\(\ell=1,\ldots,L_{\mathrm{enc}}\)}

            \State compute the backbone-layer outputs:
            \[
            \mathbf{h}_i^\ell
            \leftarrow
            \mathcal{B}_\ell
            \left(
                \left\{
                \widetilde{\mathbf{h}}_j^{\ell-1}
                \right\}_{j=0}^{N}
            \right),
            \quad i=0,\ldots,N
            \]

            \State inject the side-branch outputs:
            \[
            \widetilde{\mathbf{h}}_i^\ell
            \leftarrow
            \mathbf{h}_i^\ell
            +
            \alpha\,
            \psi_\ell
            \left(
                \mathbf{h}_i^\ell,
                \mathbf{u}_{r,k}
            \right),
            \quad i=0,\ldots,N
            \]

        \EndFor

        \State construct solutions using POMO rollouts with
        \(\widetilde{\mathbf{h}}_i^{L_{\mathrm{enc}}}\)

        \State compute the policy-gradient loss and update
        \(\theta\) and \(\alpha\)

    \EndFor

    \If{\(\mathrm{epoch}\bmod T_{\mathrm{ctrl}}=0\)
    \textbf{ and } \(\mathrm{epoch}<E\)}

        \State \(r\leftarrow r+1\)

        \For{each validation task \(k=1,\ldots,16\)}

            \State evaluate the updated solver on
            \(\mathcal{D}_k\) and obtain \(y_{r,k}\)

            \State compute \(\eta_{r,k}\) and
            \(\Delta\eta_{r,k}\), and update
            \(\mathcal{H}_{r,k}\)

        \EndFor

        \State construct the training state
        \(\mathcal{I}_r\)

        \State query the pretrained LLM:
        \[
        \boldsymbol{\gamma}_r
        \leftarrow
        \mathrm{LLM}
        \left(
            \mathcal{I}_r;\mathcal{P}
        \right)
        \]
        \Comment{applied from epoch \(\mathrm{epoch}+1\)}

    \EndIf

\EndFor

\State retain the final guidance
\(\boldsymbol{\gamma}_r\) and remove the LLM

\end{algorithmic}
\end{algorithm}

\subsection{Encoder}

Given a task with constraint vector \(\mathbf{z}_k\), the static feature of
node \(v_i\) is
\(\mathbf{f}_i=[\mathbf{x}_i,q_i,e_i,l_i]\), where features inactive for the
current task are padded to zero. A linear projection produces the initial
embedding:
\begin{equation}
    \mathbf{h}_i^0
    =
    W_{\mathrm{in}}\mathbf{f}_i,
    \qquad
    \widetilde{\mathbf{h}}_i^0
    =
    \mathbf{h}_i^0,
    \qquad
    i=0,\ldots,N.
\end{equation}

At encoder layer \(\ell\), the backbone computations receive
\(\{\widetilde{\mathbf{h}}_j^{\ell-1}\}_{j=0}^{N}\), which contains the
outputs of the preceding encoder layer after side-branch injection. The Multi-Head Attention (MHA), Instance Normalization (IN), and Feed-Forward Network (FFN)
sublayers produce the backbone-layer output \(\mathbf{h}_i^\ell\):
\begin{equation}
    \widehat{\mathbf{h}}_i^\ell
    =
    \mathrm{IN}
    \left(
        \widetilde{\mathbf{h}}_i^{\ell-1}
        +
        \mathrm{MHA}_i^\ell
        \left(
            \widetilde{\mathbf{h}}_0^{\ell-1},
            \ldots,
            \widetilde{\mathbf{h}}_N^{\ell-1}
        \right)
    \right),
\end{equation}
\begin{equation}
    \mathbf{h}_i^\ell
    =
    \mathrm{IN}
    \left(
        \widehat{\mathbf{h}}_i^\ell
        +
        \mathrm{FFN}^\ell
        \left(
            \widehat{\mathbf{h}}_i^\ell
        \right)
    \right).
    \label{eq:enc-layer}
\end{equation}
The side branch in Eq.~\ref{eq:inject} subsequently maps
\(\mathbf{h}_i^\ell\) to \(\widetilde{\mathbf{h}}_i^\ell\), which is passed
to layer \(\ell+1\). For the original backbone without LaT,
\(\widetilde{\mathbf{h}}_i^\ell=\mathbf{h}_i^\ell\).

For attention head
\(\lambda\in\{1,\ldots,H_{\mathrm{att}}\}\), the query, key, and value are
linear projections of the preceding encoder output:
\begin{equation}
    \mathbf{q}_i^{\ell,\lambda}
    =
    W_Q^{\ell,\lambda}
    \widetilde{\mathbf{h}}_i^{\ell-1},
    \qquad
    \mathbf{k}_i^{\ell,\lambda}
    =
    W_K^{\ell,\lambda}
    \widetilde{\mathbf{h}}_i^{\ell-1},
    \qquad
    \mathbf{v}_i^{\ell,\lambda}
    =
    W_V^{\ell,\lambda}
    \widetilde{\mathbf{h}}_i^{\ell-1}.
\end{equation}
The compatibility and attention weight are:
\begin{equation}
    \beta_{ij}^{\ell,\lambda}
    =
    \frac{
        (\mathbf{q}_i^{\ell,\lambda})^\top
        \mathbf{k}_j^{\ell,\lambda}
    }{
        \sqrt{d_{\mathrm{att}}}
    },
    \qquad
    \omega_{ij}^{\ell,\lambda}
    =
    \frac{
        \exp(\beta_{ij}^{\ell,\lambda})
    }{
        \sum_{j'=0}^{N}
        \exp(\beta_{ij'}^{\ell,\lambda})
    }.
\end{equation}
Each head aggregates the values over the depot and all customer nodes:
\begin{equation}
    \boldsymbol{\zeta}_i^{\ell,\lambda}
    =
    \sum_{j=0}^{N}
    \omega_{ij}^{\ell,\lambda}
    \mathbf{v}_j^{\ell,\lambda},
    \qquad
    \mathrm{MHA}_i^\ell
    =
    W_O^\ell
    \left[
        \boldsymbol{\zeta}_i^{\ell,1},
        \ldots,
        \boldsymbol{\zeta}_i^{\ell,H_{\mathrm{att}}}
    \right].
\end{equation}
The feed-forward network is:
\begin{equation}
    \mathrm{FFN}^\ell
    \left(
        \widehat{\mathbf{h}}_i^\ell
    \right)
    =
    W_2^\ell
    \mathrm{ReLU}
    \left(
        W_1^\ell
        \widehat{\mathbf{h}}_i^\ell
        +
        \mathbf{b}_1^\ell
    \right)
    +
    \mathbf{b}_2^\ell.
\end{equation}

We use \(L_{\mathrm{enc}}=6\) encoder layers,
\(H_{\mathrm{att}}=8\) attention heads,
\(d_{\mathrm{att}}=d_{\mathrm{model}}/H_{\mathrm{att}}=16\), and
\(d_{\mathrm{model}}=128\). The encoder is executed once per instance, and
the final embeddings
\(\widetilde{\mathbf{h}}_i^{L_{\mathrm{enc}}}\) are reused at every decoding
step.

\subsection{Decoder}

The decoder constructs a solution autoregressively. At step \(t\), it forms a
context embedding from the final encoder embedding of the last selected node
and the dynamic state vector:
\begin{equation}
    \mathbf{h}_{\mathrm{ctx}}
    =
    \left[
        \widetilde{\mathbf{h}}_{\tau_{t-1}}^{L_{\mathrm{enc}}},
        \mathbf{d}_t
    \right],
    \qquad
    \mathbf{d}_t
    =
    [Q_t^{\mathrm{rem}},\vartheta_t,D_t,o_t],
    \label{eq:context}
\end{equation}
where \(Q_t^{\mathrm{rem}}\) is the remaining capacity,
\(\vartheta_t\) is the current time, \(D_t\) is the accumulated route length,
and \(o_t\) is the open-route indicator.

A multi-head attention layer without self-attention computes the decoder
context. The context provides the query, while the final LaT encoder
embeddings provide the keys and values:
\begin{equation}
    \mathbf{q}_{\mathrm{ctx}}^\lambda
    =
    W_Q^{\mathrm{dec},\lambda}
    \mathbf{h}_{\mathrm{ctx}},
    \qquad
    \mathbf{k}_i^\lambda
    =
    W_K^{\mathrm{dec},\lambda}
    \widetilde{\mathbf{h}}_i^{L_{\mathrm{enc}}},
    \qquad
    \mathbf{v}_i^\lambda
    =
    W_V^{\mathrm{dec},\lambda}
    \widetilde{\mathbf{h}}_i^{L_{\mathrm{enc}}}.
\end{equation}
The same attention weighting and head concatenation used in the encoder
produce the updated decoder context
\(\widehat{\mathbf{h}}_{\mathrm{ctx}}\). The key used by the final
single-head compatibility layer is:
\begin{equation}
    \mathbf{k}_i^{\mathrm{score}}
    =
    W_K^{\mathrm{score}}
    \widetilde{\mathbf{h}}_i^{L_{\mathrm{enc}}}.
\end{equation}
The selection logits are:
\begin{equation}
    g_i
    =
    \begin{cases}
        \xi\cdot
        \tanh
        \left(
            \dfrac{
                \widehat{\mathbf{h}}_{\mathrm{ctx}}^\top
                \mathbf{k}_i^{\mathrm{score}}
            }{
                \sqrt{d_{\mathrm{att}}}
            }
        \right),
        & i\notin\mathcal{V}_t^{\mathrm{inf}},\\[6pt]
        -\infty,
        & i\in\mathcal{V}_t^{\mathrm{inf}},
    \end{cases}
\end{equation}
and the selection probability is:
\begin{equation}
    \pi_\theta
    \left(
        a_t=i
        \mid
        s_t,G,\mathbf{z}_k
    \right)
    =
    \frac{\exp(g_i)}
    {\sum_{j=0}^{N}\exp(g_j)},
\end{equation}
where \(\mathcal{V}_t^{\mathrm{inf}}\) is the set of infeasible nodes and
\(\xi=10\) is the clipping constant.

\subsection{Training Procedure}
\label{sec:pnp}
Each training batch samples one VRP variant \(k\) and generates a batch
of instances under \(\mathbf{z}_k\). The solver constructs solutions
with POMO rollouts using the final encoder embeddings
\(\widetilde{\mathbf{h}}_i^{L_{\mathrm{enc}}}\) and updates its parameters \(\theta\)
and coefficient \(\alpha\) through the original reinforcement learning
loss. Thus, LaT changes neither the decoding mechanism nor the
optimization objective. After every \(T_{\mathrm{ctrl}}\) completed
epochs, provided that additional training epochs remain, LaT evaluates
the solver on the \(16\) validation sets, computes the relative
reference gaps, assembles the training state, and queries the pretrained
LLM to update the guidance vector. The updated guidance is used from
the next training epoch. The LLM receives no gradient, is never queried
during decoding, and is invoked only
\(\lfloor(E-1)/T_{\mathrm{ctrl}}\rfloor\) times throughout training.
After training, the final guidance vector is fixed and the LLM is
removed, while the learned encoder side branches remain in the deployed
solver. Consequently, inference requires no LLM calls, and the
remaining modules introduce only negligible computational overhead
relative to the backbone. The complete procedure is summarized in
Algorithm~\ref{alg:lat}.

\section{Sensitivity to the Guidance Update Interval}
\label{app:update-interval-sensitivity}

\begin{figure}[htbp]
  \centering
  \includegraphics[width=0.4\linewidth]{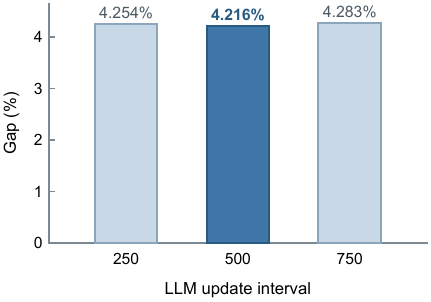}
  \caption{Sensitivity analysis of the LLM update interval on 50-node multi-task VRPs. All results are obtained with LaT on the POMO-MTL backbone.}
  \label{fig:update-interval-sensitivity}
\end{figure}

We examine the effect of the guidance-update interval, defined as the number
of training epochs between two consecutive guidance updates. We evaluate
intervals of \(250\), \(500\), and \(750\) epochs on 50-node multi-task VRPs
using the same LaT-POMO-MTL backbone, training configuration, and evaluation
protocol. For each setting, cross-task validation is performed immediately
before every guidance update. Therefore, the validation interval is identical
to \(T_{\mathrm{ctrl}}\), and each LLM call uses the latest validation results
obtained from the current solver. As shown in
Fig.~\ref{fig:update-interval-sensitivity}, the interval of \(500\) epochs
achieves the lowest average gap of \(4.216\%\), compared with \(4.254\%\) and
\(4.283\%\) for intervals of \(250\) and \(750\), respectively. It also
obtains the lowest or tied-lowest gap on most tasks. Therefore, we use
\(T_{\mathrm{ctrl}}=500\) as the default guidance-update interval.

\section{Additional Experiments}
\label{app:additional-experiments}

\subsection{Comparison with Plug-and-Play Frameworks}
\label{app:plugin-framework-comparison}

We compare LaT with MoSES and PoMtVRS, two recent state-of-the-art
plug-and-play frameworks for multi-task VRPs. MoSES constructs specialized
LoRA experts for basis VRP variants and combines their representations through
a gating mechanism \citep{moses}. PoMtVRS improves decoder representations
through a preference-gated block and optimizes the policy using pairwise
preferences between candidate solutions \citep{meng2026pomt}. Both frameworks
are integrated with CaDA to obtain MoSES-CaDA and PoMtVRS-CaDA, respectively.
All methods use the same CaDA backbone and are evaluated under identical
settings on \(16\) 50-node VRP variants.

As shown in Table~\ref{tab:plugin_framework_comparison}, all three frameworks
improve the original CaDA model. LaT-CaDA obtains an average score of 10.0497
and the lowest reported average gap of \(3.8068\%\). Its average gap is
\(0.3008\%\) lower than that of CaDA, \(0.0039\%\) lower than that of
PoMtVRS-CaDA, and \(0.2020\%\) lower than that of MoSES-CaDA. Thus, LaT-CaDA
achieves the lowest mean gap among the plug-and-play frameworks considered in
this comparison.

\begin{table}[htbp]
  \centering
  \caption{Comparison with recent state-of-the-art plug-and-play
  frameworks using CaDA on 16 50-node VRP variants.}
  \label{tab:plugin_framework_comparison}
  \setlength{\tabcolsep}{6pt}
  \renewcommand{\arraystretch}{1.05}
  \begin{tabular}{@{}lcc@{}}
    \toprule
    Method
    & Avg. Score $\downarrow$
    & Avg. Gap $\downarrow$
    \\
    \midrule
    CaDA
    & 10.0795
    & 4.1076\%
    \\
    MoSES-CaDA
    & 10.0643
    & 4.0088\%
    \\
    PoMtVRS-CaDA
    & 10.0524
    & 3.8107\%
    \\
    LaT-CaDA
    & \textbf{10.0497}
    & \textbf{3.8068\%}
    \\
    \bottomrule
  \end{tabular}
\end{table}

\subsection{Performance across Backbones and VRP Variants}
\label{app:backbone-results}
To evaluate whether LaT can improve neural solvers with different model architectures, we integrated it with POMO-MTL, MVMoE-L, and MVMoE. Each LaT variant was compared with its original backbone and the corresponding Mixed variant \citep{liu2025mixed} under the same problem size.

For 50-node instances, Table~\ref{crossProblem} shows that LaT reduced the average gap of POMO-MTL from 4.602\% to 4.216\%. It also reduced the average gaps of MVMoE-L and MVMoE from 4.522\% to 4.081\% and from 4.352\% to 3.980\%, respectively. The corresponding Mixed variants achieved average gaps of 4.603\%, 4.416\%, and 4.279\%. All three LaT variants therefore outperformed both their original backbones and the corresponding Mixed variants. Among the evaluated models, LaT-MVMoE achieved the lowest average gap of 3.980\%.

\begin{table*}[!t]
  \centering
  \vspace{-0.05in}
  \caption{Performance on 1K test instances of 16 VRP variants with
  $N=50$. Blue rows denote our LaT variants. The best results are
  shown in bold.}
  \label{crossProblem}
  \vspace{0.05in}

  \begin{small}
  \setlength{\tabcolsep}{3.0pt}
  \renewcommand{\arraystretch}{0.60}

  \resizebox{0.98\textwidth}{!}{%
  \begin{tabular}{@{}l*{4}{rrr}@{}}
    \toprule
    & \multicolumn{3}{c}{CVRP}
    & \multicolumn{3}{c}{VRPTW}
    & \multicolumn{3}{c}{OVRP}
    & \multicolumn{3}{c}{VRPL}
    \\
    \cmidrule(lr){2-4}
    \cmidrule(lr){5-7}
    \cmidrule(lr){8-10}
    \cmidrule(lr){11-13}
    Method
    & Obj. & Gap & Time
    & Obj. & Gap & Time
    & Obj. & Gap & Time
    & Obj. & Gap & Time
    \\
    \midrule
    HGS
    & 10.334 & 0.000\% & 4.6m
    & 14.509 & 0.000\% & 8.4m
    & -- & -- & --
    & -- & -- & --
    \\
    LKH3
    & 10.346 & 0.115\% & 9.9m
    & 14.607 & 0.664\% & 5.5m
    & 6.511 & 0.198\% & 4.5m
    & 10.571 & 0.790\% & 7.8m
    \\
    OR-Tools
    & 10.540 & 1.962\% & 10.4m
    & 14.915 & 2.694\% & 10.4m
    & 6.531 & 0.495\% & 10.4m
    & 10.677 & 1.746\% & 10.4m
    \\
    OR-Tools ($\times 10$)
    & 10.418 & 0.788\% & 1.7h
    & 14.665 & 1.011\% & 1.7h
    & 6.498 & 0.000\% & 1.7h
    & 10.495 & 0.000\% & 1.7h
    \\
    \midrule
    POMO-MTL
    & 10.437 & 0.987\% & 3s
    & 15.032 & 3.637\% & 3s
    & 6.671 & 2.634\% & 2s
    & 10.513 & 0.201\% & 2s
    \\
    Mixed-POMO-MTL
    & 10.436 & 0.980\% & 5s
    & 15.021 & 3.556\% & 4s
    & 6.670 & 2.637\% & 3s
    & 10.511 & 0.185\% & 3s
    \\
    \rowcolor{latrow}
    LaT-POMO-MTL
    & 10.431 & 0.932\% & 3s
    & 14.992 & 3.350\% & 3s
    & 6.639 & 2.144\% & 2s
    & 10.505 & 0.132\% & 2s
    \\
    \familyrule
    MVMoE-L
    & 10.434 & 0.955\% & 4s
    & 15.013 & 3.500\% & 4s
    & 6.665 & 2.548\% & 3s
    & 10.506 & 0.131\% & 3s
    \\
    Mixed-MVMoE-L
    & 10.431 & 0.933\% & 6s
    & 15.002 & 3.421\% & 4s
    & 6.658 & 2.448\% & 4s
    & 10.502 & 0.098\% & 3s
    \\
    \rowcolor{latrow}
    LaT-MVMoE-L
    & 10.427 & 0.884\% & 4s
    & 14.975 & 3.240\% & 4s
    & 6.636 & 2.105\% & 3s
    & 10.499 & 0.067\% & 3s
    \\
    \familyrule
    MVMoE
    & 10.428 & 0.896\% & 4s
    & 14.999 & 3.410\% & 4s
    & 6.655 & 2.402\% & 3s
    & 10.501 & 0.092\% & 3s
    \\
    Mixed-MVMoE
    & 10.424 & 0.865\% & 7s
    & 14.995 & 3.373\% & 4s
    & 6.651 & 2.336\% & 4s
    & 10.497 & 0.052\% & 4s
    \\
    \rowcolor{latrow}
    LaT-MVMoE
    & \textbf{10.421} & \textbf{0.834\%} & 4s
    & \textbf{14.965} & \textbf{3.154\%} & 4s
    & \textbf{6.622} & \textbf{1.888\%} & 3s
    & \textbf{10.494} & \textbf{0.015\%} & 3s
    \\

    \midrule
    & \multicolumn{3}{c}{VRPB}
    & \multicolumn{3}{c}{OVRPTW}
    & \multicolumn{3}{c}{OVRPB}
    & \multicolumn{3}{c}{OVRPL}
    \\
    \cmidrule(lr){2-4}
    \cmidrule(lr){5-7}
    \cmidrule(lr){8-10}
    \cmidrule(lr){11-13}
    Method
    & Obj. & Gap & Time
    & Obj. & Gap & Time
    & Obj. & Gap & Time
    & Obj. & Gap & Time
    \\
    \midrule
    OR-Tools
    & 8.127 & 0.989\% & 10.4m
    & 8.737 & 0.592\% & 10.4m
    & 5.764 & 0.332\% & 10.4m
    & 6.522 & 0.480\% & 10.4m
    \\
    OR-Tools ($\times 10$)
    & 8.046 & 0.000\% & 1.7h
    & 8.683 & 0.000\% & 1.7h
    & 5.745 & 0.000\% & 1.7h
    & 6.490 & 0.000\% & 1.7h
    \\
    \midrule
    POMO-MTL
    & 8.182 & 1.684\% & 2s
    & 8.987 & 3.470\% & 3s
    & 6.116 & 6.430\% & 2s
    & 6.668 & 2.734\% & 2s
    \\
    Mixed-POMO-MTL
    & 8.179 & 1.645\% & 2s
    & 8.982 & 3.420\% & 3s
    & 6.112 & 6.348\% & 3s
    & 6.667 & 2.708\% & 3s
    \\
    \rowcolor{latrow}
    LaT-POMO-MTL
    & 8.175 & 1.598\% & 2s
    & 8.933 & 2.853\% & 3s
    & 6.057 & 5.398\% & 2s
    & 6.637 & 2.249\% & 2s
    \\
    \familyrule
    MVMoE-L
    & 8.176 & 1.605\% & 3s
    & 8.974 & 3.322\% & 4s
    & 6.122 & 6.522\% & 3s
    & 6.659 & 2.597\% & 3s
    \\
    Mixed-MVMoE-L
    & 8.170 & 1.531\% & 3s
    & 8.964 & 3.219\% & 4s
    & 6.102 & 6.175\% & 3s
    & 6.653 & 2.497\% & 4s
    \\
    \rowcolor{latrow}
    LaT-MVMoE-L
    & 8.169 & 1.527\% & 3s
    & 8.925 & 2.760\% & 4s
    & 6.047 & 5.225\% & 3s
    & 6.632 & 2.175\% & 3s
    \\
    \familyrule
    MVMoE
    & 8.170 & 1.540\% & 3s
    & 8.964 & 3.210\% & 4s
    & 6.092 & 5.999\% & 3s
    & 6.650 & 2.454\% & 3s
    \\
    Mixed-MVMoE
    & 8.164 & 1.456\% & 3s
    & 8.950 & 3.060\% & 4s
    & 6.084 & 5.871\% & 4s
    & 6.648 & 2.419\% & 4s
    \\
    \rowcolor{latrow}
    LaT-MVMoE
    & \textbf{8.160} & \textbf{1.409\%} & 3s
    & \textbf{8.917} & \textbf{2.677\%} & 4s
    & \textbf{6.044} & \textbf{5.171\%} & 3s
    & \textbf{6.619} & \textbf{1.969\%} & 3s
    \\

    \midrule
    & \multicolumn{3}{c}{VRPBL}
    & \multicolumn{3}{c}{VRPBTW}
    & \multicolumn{3}{c}{VRPLTW}
    & \multicolumn{3}{c}{OVRPBL}
    \\
    \cmidrule(lr){2-4}
    \cmidrule(lr){5-7}
    \cmidrule(lr){8-10}
    \cmidrule(lr){11-13}
    Method
    & Obj. & Gap & Time
    & Obj. & Gap & Time
    & Obj. & Gap & Time
    & Obj. & Gap & Time
    \\
    \midrule
    OR-Tools
    & 8.131 & 1.254\% & 10.4m
    & 15.053 & 1.857\% & 10.4m
    & 14.815 & 1.432\% & 10.4m
    & 5.771 & 0.549\% & 10.4m
    \\
    OR-Tools ($\times 10$)
    & 8.029 & 0.000\% & 1.7h
    & 14.771 & 0.000\% & 1.7h
    & 14.598 & 0.000\% & 1.7h
    & 5.739 & 0.000\% & 1.7h
    \\
    \midrule
    POMO-MTL
    & 8.188 & 1.971\% & 2s
    & 16.055 & 8.841\% & 3s
    & 14.961 & 2.586\% & 3s
    & 6.104 & 6.306\% & 2s
    \\
    Mixed-POMO-MTL
    & 8.182 & 1.905\% & 3s
    & 16.071 & 8.943\% & 3s
    & 14.966 & 2.621\% & 3s
    & 6.102 & 6.282\% & 3s
    \\
    \rowcolor{latrow}
    LaT-POMO-MTL
    & 8.183 & 1.914\% & 2s
    & 16.049 & 8.786\% & 3s
    & 14.921 & 2.311\% & 2s
    & 6.042 & 5.242\% & 2s
    \\
    \familyrule
    MVMoE-L
    & 8.180 & 1.872\% & 3s
    & 16.041 & 8.745\% & 3s
    & 14.953 & 2.535\% & 4s
    & 6.104 & 6.310\% & 3s
    \\
    Mixed-MVMoE-L
    & 8.172 & 1.781\% & 3s
    & 16.039 & 8.715\% & 4s
    & 14.941 & 2.448\% & 4s
    & 6.090 & 6.077\% & 3s
    \\
    \rowcolor{latrow}
    LaT-MVMoE-L
    & 8.173 & 1.789\% & 3s
    & 15.996 & 8.443\% & 3s
    & 14.919 & 2.296\% & 4s
    & 6.033 & 5.084\% & 3s
    \\
    \familyrule
    MVMoE
    & 8.172 & 1.776\% & 3s
    & 16.022 & 8.600\% & 3s
    & 14.937 & 2.421\% & 4s
    & 6.076 & 5.843\% & 3s
    \\
    Mixed-MVMoE
    & 8.168 & 1.729\% & 4s
    & 16.014 & 8.545\% & 4s
    & 14.931 & 2.387\% & 4s
    & 6.068 & 5.705\% & 4s
    \\
    \rowcolor{latrow}
    LaT-MVMoE
    & \textbf{8.166} & \textbf{1.700\%} & 3s
    & \textbf{15.990} & \textbf{8.376\%} & 3s
    & \textbf{14.901} & \textbf{2.170\%} & 4s
    & \textbf{6.030} & \textbf{5.026\%} & 3s
    \\

    \midrule
    & \multicolumn{3}{c}{OVRPBTW}
    & \multicolumn{3}{c}{OVRPLTW}
    & \multicolumn{3}{c}{VRPBLTW}
    & \multicolumn{3}{c}{OVRPBLTW}
    \\
    \cmidrule(lr){2-4}
    \cmidrule(lr){5-7}
    \cmidrule(lr){8-10}
    \cmidrule(lr){11-13}
    Method
    & Obj. & Gap & Time
    & Obj. & Gap & Time
    & Obj. & Gap & Time
    & Obj. & Gap & Time
    \\
    \midrule
    OR-Tools
    & 8.758 & 0.927\% & 10.4m
    & 8.728 & 0.656\% & 10.4m
    & 14.890 & 1.402\% & 10.4m
    & 8.729 & 0.624\% & 10.4m
    \\
    OR-Tools ($\times 10$)
    & 8.675 & 0.000\% & 1.7h
    & 8.669 & 0.000\% & 1.7h
    & 14.677 & 0.000\% & 1.7h
    & 8.673 & 0.000\% & 1.7h
    \\
    \midrule
    POMO-MTL
    & 9.514 & 9.628\% & 3s
    & 8.987 & 3.633\% & 3s
    & 15.980 & 9.035\% & 3s
    & 9.532 & 9.851\% & 3s
    \\
    Mixed-POMO-MTL
    & 9.523 & 9.734\% & 3s
    & 8.984 & 3.600\% & 3s
    & 15.998 & 9.139\% & 3s
    & 9.541 & 9.946\% & 3s
    \\
    \rowcolor{latrow}
    LaT-POMO-MTL
    & 9.469 & 9.115\% & 3s
    & 8.931 & 2.993\% & 3s
    & 15.984 & 9.026\% & 3s
    & 9.493 & 9.406\% & 3s
    \\
    \familyrule
    MVMoE-L
    & 9.515 & 9.630\% & 3s
    & 8.974 & 3.488\% & 4s
    & 15.963 & 8.915\% & 4s
    & 9.518 & 9.682\% & 4s
    \\
    Mixed-MVMoE-L
    & 9.506 & 9.530\% & 4s
    & 8.961 & 3.335\% & 4s
    & 15.961 & 8.871\% & 4s
    & 9.509 & 9.582\% & 4s
    \\
    \rowcolor{latrow}
    LaT-MVMoE-L
    & 9.450 & 8.895\% & 3s
    & 8.930 & 2.975\% & 4s
    & 15.929 & 8.664\% & 4s
    & 9.472 & 9.173\% & 4s
    \\
    \familyrule
    MVMoE
    & 9.486 & 9.308\% & 4s
    & 8.966 & 3.396\% & 4s
    & 15.945 & 8.775\% & 4s
    & 9.503 & 9.516\% & 4s
    \\
    Mixed-MVMoE
    & 9.483 & 9.283\% & 4s
    & 8.951 & 3.225\% & 4s
    & 15.932 & 8.690\% & 4s
    & 9.498 & 9.462\% & 4s
    \\
    \rowcolor{latrow}
    LaT-MVMoE
    & \textbf{9.445} & \textbf{8.837\%} & 4s
    & \textbf{8.918} & \textbf{2.835\%} & 4s
    & \textbf{15.914} & \textbf{8.558\%} & 4s
    & \textbf{9.463} & \textbf{9.057\%} & 4s
    \\
    \bottomrule
  \end{tabular}%
  }
  \end{small}
  \vspace{-0.10in}
\end{table*}

For 100-node instances, Table~\ref{crossProblemN100} shows that LaT reduced the average gap of POMO-MTL from 4.969\% to 4.602\%. The average gaps of MVMoE-L and MVMoE also decreased from 4.712\% to 4.273\% and from 4.596\% to 4.170\%, respectively. The corresponding Mixed variants obtained average gaps of 4.868\%, 4.566\%, and 4.428\%, all of which were higher than those of the LaT variants. LaT-MVMoE achieved the lowest average gap of 4.170\% and obtained the best individual results on most VRP variants. Across both problem sizes, LaT reduced the average gaps of all three evaluated
backbones and obtained lower average gaps than the corresponding Mixed
variants.

\begin{table*}[!t]
  \centering
  \vspace{-0.05in}
  \caption{Performance on 1K test instances of 16 VRP variants with
  $N=100$. Blue rows denote our LaT variants. The best results are
  shown in bold.}
  \label{crossProblemN100}
  \vspace{0.05in}

  \begin{small}
  \setlength{\tabcolsep}{3.0pt}
  \renewcommand{\arraystretch}{0.55}

  \resizebox{0.98\textwidth}{!}{%
  \begin{tabular}{@{}l*{4}{rrr}@{}}
    \toprule
    & \multicolumn{3}{c}{CVRP}
    & \multicolumn{3}{c}{VRPTW}
    & \multicolumn{3}{c}{OVRP}
    & \multicolumn{3}{c}{VRPL}
    \\
    \cmidrule(lr){2-4}
    \cmidrule(lr){5-7}
    \cmidrule(lr){8-10}
    \cmidrule(lr){11-13}
    Method
    & Obj. & Gap & Time
    & Obj. & Gap & Time
    & Obj. & Gap & Time
    & Obj. & Gap & Time
    \\
    \midrule
    HGS
    & 15.504 & 0.000\% & 9.1m
    & 24.339 & 0.000\% & 19.6m
    & -- & -- & --
    & -- & -- & --
    \\
    LKH3
    & 15.590 & 0.556\% & 18.0m
    & 24.721 & 1.584\% & 7.8m
    & 9.828 & 0.000\% & 5.3m
    & 15.771 & 0.000\% & 16.0m
    \\
    OR-Tools
    & 16.381 & 5.652\% & 20.8m
    & 25.894 & 6.297\% & 20.8m
    & 10.010 & 1.806\% & 20.8m
    & 16.496 & 4.587\% & 20.8m
    \\
    OR-Tools ($\times 10$)
    & 15.935 & 2.751\% & 3.5h
    & 25.212 & 3.482\% & 3.5h
    & 9.842 & 0.122\% & 3.5h
    & 16.004 & 1.444\% & 3.5h
    \\
    \midrule
    POMO-MTL
    & 15.790 & 1.846\% & 9s
    & 25.610 & 5.313\% & 12s
    & 10.169 & 3.458\% & 9s
    & 15.846 & 0.479\% & 10s
    \\
    Mixed-POMO-MTL
    & 15.771 & 1.731\% & 14s
    & 25.556 & 5.090\% & 12s
    & 10.154 & 3.312\% & 10s
    & 15.827 & 0.362\% & 11s
    \\
    \rowcolor{latrow}
    LaT-POMO-MTL
    & 15.775 & 1.751\% & 9s
    & 25.520 & 4.933\% & 12s
    & 10.123 & 2.991\% & 9s
    & 15.830 & 0.379\% & 10s
    \\
    \familyrule
    MVMoE-L
    & 15.771 & 1.728\% & 11s
    & 25.519 & 4.927\% & 14s
    & 10.145 & 3.214\% & 11s
    & 15.821 & 0.323\% & 12s
    \\
    Mixed-MVMoE-L
    & 15.758 & 1.645\% & 14s
    & 25.506 & 4.872\% & 15s
    & 10.136 & 3.133\% & 12s
    & 15.813 & 0.270\% & 13s
    \\
    \rowcolor{latrow}
    LaT-MVMoE-L
    & 15.755 & 1.622\% & 11s
    & 25.451 & 4.639\% & 14s
    & 10.096 & 2.717\% & 11s
    & 15.810 & 0.255\% & 12s
    \\
    \familyrule
    MVMoE
    & 15.760 & 1.653\% & 12s
    & 25.512 & 4.903\% & 15s
    & 10.138 & 3.136\% & 12s
    & 15.812 & 0.261\% & 14s
    \\
    Mixed-MVMoE
    & 15.751 & 1.599\% & 16s
    & 25.473 & 4.732\% & 16s
    & 10.119 & 2.946\% & 12s
    & 15.806 & 0.227\% & 14s
    \\
    \rowcolor{latrow}
    LaT-MVMoE
    & \textbf{15.743} & \textbf{1.545\%} & 12s
    & \textbf{25.436} & \textbf{4.578\%} & 15s
    & \textbf{10.083} & \textbf{2.584\%} & 12s
    & \textbf{15.800} & \textbf{0.192\%} & 14s
    \\

    \midrule
    & \multicolumn{3}{c}{VRPB}
    & \multicolumn{3}{c}{OVRPTW}
    & \multicolumn{3}{c}{OVRPB}
    & \multicolumn{3}{c}{OVRPL}
    \\
    \cmidrule(lr){2-4}
    \cmidrule(lr){5-7}
    \cmidrule(lr){8-10}
    \cmidrule(lr){11-13}
    Method
    & Obj. & Gap & Time
    & Obj. & Gap & Time
    & Obj. & Gap & Time
    & Obj. & Gap & Time
    \\
    \midrule
    OR-Tools
    & 12.185 & 2.594\% & 20.8m
    & 14.635 & 1.756\% & 20.8m
    & 8.522 & 1.852\% & 20.8m
    & 9.966 & 1.783\% & 20.8m
    \\
    OR-Tools ($\times 10$)
    & 11.878 & 0.000\% & 3.5h
    & 14.380 & 0.000\% & 3.5h
    & 8.365 & 0.000\% & 3.5h
    & 9.790 & 0.000\% & 3.5h
    \\
    \midrule
    POMO-MTL
    & 12.072 & 1.674\% & 8s
    & 15.008 & 4.411\% & 12s
    & 8.979 & 7.335\% & 8s
    & 10.126 & 3.441\% & 10s
    \\
    Mixed-POMO-MTL
    & 12.043 & 1.427\% & 8s
    & 14.948 & 3.996\% & 12s
    & 9.021 & 7.831\% & 9s
    & 10.116 & 3.350\% & 11s
    \\
    \rowcolor{latrow}
    LaT-POMO-MTL
    & 12.057 & 1.539\% & 8s
    & 14.905 & 3.707\% & 12s
    & 8.913 & 6.546\% & 8s
    & 10.082 & 2.996\% & 10s
    \\
    \familyrule
    MVMoE-L
    & 12.036 & 1.368\% & 10s
    & 14.940 & 3.941\% & 14s
    & 8.972 & 7.243\% & 10s
    & 10.106 & 3.244\% & 12s
    \\
    Mixed-MVMoE-L
    & 12.025 & 1.265\% & 10s
    & 14.911 & 3.749\% & 15s
    & 8.951 & 6.997\% & 11s
    & 10.098 & 3.159\% & 13s
    \\
    \rowcolor{latrow}
    LaT-MVMoE-L
    & 12.033 & 1.333\% & 10s
    & 14.841 & 3.259\% & 14s
    & 8.885 & 6.199\% & 10s
    & 10.052 & 2.694\% & 12s
    \\
    \familyrule
    MVMoE
    & 12.027 & 1.285\% & 10s
    & 14.927 & 3.852\% & 15s
    & 8.959 & 7.088\% & 11s
    & 10.097 & 3.148\% & 13s
    \\
    Mixed-MVMoE
    & \textbf{12.011} & \textbf{1.153\%} & 11s
    & 14.888 & 3.579\% & 16s
    & 8.934 & 6.800\% & 12s
    & 10.079 & 2.971\% & 14s
    \\
    \rowcolor{latrow}
    LaT-MVMoE
    & 12.012 & 1.164\% & 10s
    & \textbf{14.825} & \textbf{3.145\%} & 15s
    & \textbf{8.870} & \textbf{6.026\%} & 11s
    & \textbf{10.039} & \textbf{2.560\%} & 13s
    \\

    \midrule
    & \multicolumn{3}{c}{VRPBL}
    & \multicolumn{3}{c}{VRPBTW}
    & \multicolumn{3}{c}{VRPLTW}
    & \multicolumn{3}{c}{OVRPBL}
    \\
    \cmidrule(lr){2-4}
    \cmidrule(lr){5-7}
    \cmidrule(lr){8-10}
    \cmidrule(lr){11-13}
    Method
    & Obj. & Gap & Time
    & Obj. & Gap & Time
    & Obj. & Gap & Time
    & Obj. & Gap & Time
    \\
    \midrule
    OR-Tools
    & 12.095 & 2.586\% & 20.8m
    & 26.217 & 2.858\% & 20.8m
    & 25.823 & 2.534\% & 20.8m
    & 8.555 & 2.459\% & 20.8m
    \\
    OR-Tools ($\times 10$)
    & 11.790 & 0.000\% & 3.5h
    & 25.496 & 0.000\% & 3.5h
    & 25.195 & 0.000\% & 3.5h
    & 8.348 & 0.000\% & 3.5h
    \\
    \midrule
    POMO-MTL
    & 11.998 & 1.793\% & 9s
    & 27.319 & 7.413\% & 11s
    & 25.619 & 1.920\% & 13s
    & 8.961 & 7.343\% & 9s
    \\
    Mixed-POMO-MTL
    & 11.964 & 1.514\% & 10s
    & 27.327 & 7.457\% & 12s
    & 25.561 & 1.673\% & 14s
    & 9.009 & 7.919\% & 10s
    \\
    \rowcolor{latrow}
    LaT-POMO-MTL
    & 11.985 & 1.686\% & 9s
    & 27.315 & 7.407\% & 11s
    & 25.538 & 1.597\% & 13s
    & 8.906 & 6.687\% & 9s
    \\
    \familyrule
    MVMoE-L
    & 11.960 & 1.473\% & 10s
    & 27.265 & 7.190\% & 13s
    & 25.529 & 1.545\% & 16s
    & 8.957 & 7.300\% & 11s
    \\
    Mixed-MVMoE-L
    & 11.949 & 1.378\% & 12s
    & 27.223 & 7.018\% & 11s
    & 25.521 & 1.515\% & 17s
    & 8.935 & 7.027\% & 11s
    \\
    \rowcolor{latrow}
    LaT-MVMoE-L
    & 11.952 & 1.409\% & 10s
    & 27.197 & 6.940\% & 13s
    & 25.456 & 1.247\% & 16s
    & 8.868 & 6.224\% & 11s
    \\
    \familyrule
    MVMoE
    & 11.945 & 1.346\% & 11s
    & 27.236 & 7.078\% & 14s
    & 25.514 & 1.471\% & 17s
    & 8.942 & 7.115\% & 12s
    \\
    Mixed-MVMoE
    & \textbf{11.936} & \textbf{1.264\%} & 12s
    & 27.208 & 6.967\% & 15s
    & 25.486 & 1.365\% & 18s
    & 8.920 & 6.857\% & 12s
    \\
    \rowcolor{latrow}
    LaT-MVMoE
    & 11.938 & 1.285\% & 11s
    & \textbf{27.181} & \textbf{6.862\%} & 14s
    & \textbf{25.437} & \textbf{1.172\%} & 17s
    & \textbf{8.858} & \textbf{6.102\%} & 12s
    \\

    \midrule
    & \multicolumn{3}{c}{OVRPBTW}
    & \multicolumn{3}{c}{OVRPLTW}
    & \multicolumn{3}{c}{VRPBLTW}
    & \multicolumn{3}{c}{OVRPBLTW}
    \\
    \cmidrule(lr){2-4}
    \cmidrule(lr){5-7}
    \cmidrule(lr){8-10}
    \cmidrule(lr){11-13}
    Method
    & Obj. & Gap & Time
    & Obj. & Gap & Time
    & Obj. & Gap & Time
    & Obj. & Gap & Time
    \\
    \midrule
    OR-Tools
    & 14.713 & 2.268\% & 20.8m
    & 14.535 & 1.779\% & 20.8m
    & 25.979 & 2.518\% & 20.8m
    & 14.496 & 1.724\% & 20.8m
    \\
    OR-Tools ($\times 10$)
    & 14.384 & 0.000\% & 3.5h
    & 14.279 & 0.000\% & 3.5h
    & 25.342 & 0.000\% & 3.5h
    & 14.250 & 0.000\% & 3.5h
    \\
    \midrule
    POMO-MTL
    & 15.879 & 10.453\% & 10s
    & 14.896 & 4.374\% & 12s
    & 27.247 & 7.746\% & 12s
    & 15.738 & 10.498\% & 11s
    \\
    Mixed-POMO-MTL
    & 15.844 & 10.192\% & 11s
    & 14.845 & 4.020\% & 12s
    & 27.219 & 7.658\% & 13s
    & 15.720 & 10.358\% & 13s
    \\
    \rowcolor{latrow}
    LaT-POMO-MTL
    & 15.797 & 9.877\% & 10s
    & 14.814 & 3.798\% & 12s
    & 27.222 & 7.658\% & 12s
    & 15.678 & 10.072\% & 11s
    \\
    \familyrule
    MVMoE-L
    & 15.841 & 10.188\% & 12s
    & 14.839 & 3.971\% & 14s
    & 27.177 & 7.473\% & 14s
    & 15.706 & 10.263\% & 13s
    \\
    Mixed-MVMoE-L
    & 15.802 & 9.899\% & 13s
    & 14.816 & 3.816\% & 15s
    & 27.129 & 7.278\% & 15s
    & 15.673 & 10.027\% & 14s
    \\
    \rowcolor{latrow}
    LaT-MVMoE-L
    & 15.749 & 9.537\% & 12s
    & 14.744 & 3.311\% & 14s
    & 27.129 & 7.280\% & 14s
    & 15.624 & 9.705\% & 13s
    \\
    \familyrule
    MVMoE
    & 15.808 & 9.948\% & 13s
    & 14.828 & 3.903\% & 15s
    & 27.142 & 7.332\% & 15s
    & 15.671 & 10.009\% & 14s
    \\
    Mixed-MVMoE
    & 15.779 & 9.749\% & 14s
    & 14.779 & 3.560\% & 16s
    & 27.136 & 7.304\% & 16s
    & 15.636 & 9.772\% & 16s
    \\
    \rowcolor{latrow}
    LaT-MVMoE
    & \textbf{15.742} & \textbf{9.491\%} & 13s
    & \textbf{14.726} & \textbf{3.187\%} & 15s
    & \textbf{27.107} & \textbf{7.196\%} & 15s
    & \textbf{15.616} & \textbf{9.635\%} & 14s
    \\
    \bottomrule
  \end{tabular}%
  }
  \end{small}
  \vspace{-0.10in}
\end{table*}





\subsection{Convergence Analysis}

\paragraph{Overall convergence of LaT-POMO-MTL.}
We compare the convergence of LaT-POMO-MTL with its POMO-MTL backbone over
5,000 training epochs on 16 \(N=50\) VRP variants. Both models were evaluated
every 50 epochs on the same 1,000 fixed validation instances for each variant
using eight-fold augmentation. We first computed the relative reference gap
for each variant and then averaged the gaps across all 16 variants.

\begin{figure}[htbp]
    \centering
    \includegraphics[width=0.55\linewidth]
    {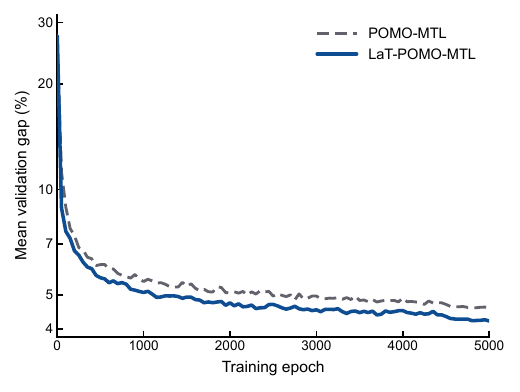}
    \caption{Mean relative reference gap during training on 16
    \(N=50\) VRP variants. The curves compare POMO-MTL and LaT-POMO-MTL.
    Lower values indicate better performance.}
    \label{fig:n50-suite-convergence}
\end{figure}

As shown in Fig.~\ref{fig:n50-suite-convergence}, LaT-POMO-MTL reduced the mean
gap faster and maintained a lower gap than POMO-MTL. LaT-POMO-MTL reached a
mean gap of \(5\%\) at epoch 1,150, whereas POMO-MTL reached the same value at
epoch 2,200. LaT-POMO-MTL also reduced the final mean gap from \(4.602\%\) to
\(4.216\%\). Its convergence curve showed smaller fluctuations during
training. These results support the effectiveness of LaT in improving the
training of POMO-MTL.

\paragraph{Convergence on difficult time-window variants.}
We further compare LaT-POMO-MTL with POMO-MTL on three difficult \(N=50\)
time-window variants, including VRPTW, OVRPBTW, and OVRPBLTW. Both models were
evaluated using the same fixed validation instances, 50-epoch evaluation
interval, and eight-fold augmentation.

\begin{figure*}[!t]
    \centering
    \begin{minipage}[t]{0.31\textwidth}
        \centering
        \includegraphics[width=\linewidth]
        {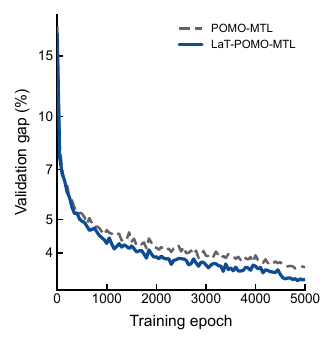}
        \par\smallskip
        \textbf{(a)} VRPTW
    \end{minipage}
    \hfill
    \begin{minipage}[t]{0.31\textwidth}
        \centering
        \includegraphics[width=\linewidth]
        {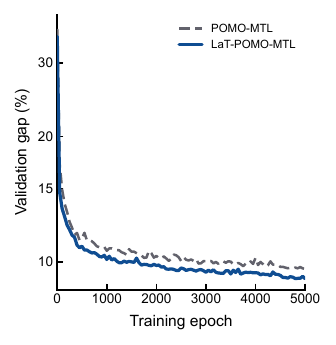}
        \par\smallskip
        \textbf{(b)} OVRPBTW
    \end{minipage}
    \hfill
    \begin{minipage}[t]{0.31\textwidth}
        \centering
        \includegraphics[width=\linewidth]
        {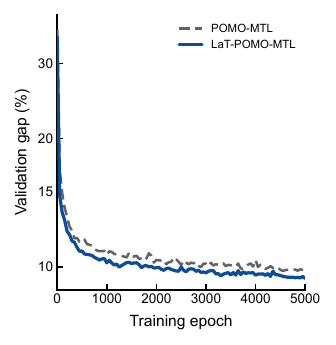}
        \par\smallskip
        \textbf{(c)} OVRPBLTW
    \end{minipage}
    \caption{Relative reference gaps during training on three
    difficult \(N=50\) time-window variants. Results are shown for
    (a) VRPTW, (b) OVRPBTW, and (c) OVRPBLTW. The curves compare POMO-MTL
    and LaT-POMO-MTL. Lower values indicate better performance. }
    \label{fig:n50-hard-task-convergence}
\end{figure*}

As shown in Fig.~\ref{fig:n50-hard-task-convergence}, LaT-POMO-MTL achieved
lower final gaps on all three variants. The final gap decreased from
\(3.637\%\) to \(3.350\%\) on VRPTW, from \(9.628\%\) to \(9.115\%\) on
OVRPBTW, and from \(9.851\%\) to \(9.406\%\) on OVRPBLTW. The largest absolute
reduction was \(0.513\) percentage points on OVRPBTW. These results further
support the effectiveness of LaT-POMO-MTL on difficult time-window variants.

\subsection{Evaluation of Performance on Benchmark Datasets}
\label{app:benchmark-results}

To evaluate out-of-distribution generalization to different problem sizes, we
test models trained on uniformly distributed instances with \(N=100\) on
selected CVRPLIB Set-X instances. As reported in
Table~\ref{tab:cvrplib_setx_lat}, LaT-POMO-MTL achieves the lowest average gap
of \(5.198\%\), compared with \(9.820\%\) for POMO-MTL and \(16.629\%\) for
POMO. LaT-MVMoE also reduces the average gap of MVMoE from \(6.884\%\) to
\(5.894\%\). The reductions obtained with both backbones support the
generalization of LaT to instances whose problem sizes differ from those used
during training.

\begin{table*}[!t]
    \centering
\caption{Comparison of neural solvers on selected CVRPLIB Set-X
instances. All models are trained on uniformly distributed instances
with \(N=100\) and evaluated using the same eight-fold instance
augmentation. The lowest objective value and its corresponding gap
for each instance are highlighted in \textbf{bold}.}
    \label{tab:cvrplib_setx_lat}
    \setlength{\tabcolsep}{3.2pt}
    \renewcommand{\arraystretch}{1.05}
    \resizebox{\textwidth}{!}{%
    \begin{tabular}{lrrrrrrrrrrr}
        \toprule
       \multirow{2}{*}{Instance}
        & \multirow{2}{*}{Opt.}
        & \multicolumn{2}{c}{POMO}
        & \multicolumn{2}{c}{POMO-MTL}
        & \multicolumn{2}{c}{LaT-POMO-MTL}
        & \multicolumn{2}{c}{MVMoE}
        & \multicolumn{2}{c}{LaT-MVMoE} \\
        \cmidrule(lr){3-4}
        \cmidrule(lr){5-6}
        \cmidrule(lr){7-8}
        \cmidrule(lr){9-10}
        \cmidrule(lr){11-12}
        & & Obj. & Gap
          & Obj. & Gap
          & Obj. & Gap
          & Obj. & Gap
          & Obj. & Gap \\
        \midrule

        X-n101-k25
        & 27591
        & 30138 & 9.231\%
        & 32482 & 17.727\%
        & 28924 & 4.831\%
        & 29361 & 6.415\%
        & \textbf{28812} & \textbf{4.425\%} \\

        X-n106-k14
        & 26362
        & 39322 & 49.162\%
        & 27369 & 3.820\%
        & 27312 & 3.604\%
        & \textbf{27278} & \textbf{3.475\%}
        & 27354 & 3.763\% \\

        X-n110-k13
        & 14971
        & 15223 & 1.683\%
        & 15151 & 1.202\%
        & 15274 & 2.024\%
        & 15089 & 0.788\%
        & \textbf{15040} & \textbf{0.461\%} \\

        X-n115-k10
        & 12747
        & 16113 & 26.406\%
        & 14785 & 15.988\%
        & \textbf{13204} & \textbf{3.585\%}
        & 13847 & 8.629\%
        & 13468 & 5.656\% \\

        X-n120-k6
        & 13332
        & 14085 & 5.648\%
        & 13931 & 4.493\%
        & 13634 & 2.265\%
        & 14089 & 5.678\%
        & \textbf{13633} & \textbf{2.258\%} \\

        X-n125-k30
        & 55539
        & 58513 & 5.355\%
        & 60687 & 9.269\%
        & \textbf{58348} & \textbf{5.058\%}
        & 58944 & 6.131\%
        & 59078 & 6.372\% \\

         X-n129-k18
        & 28940
        & \textbf{29246} & \textbf{1.057\%}
        & 30332 & 4.810\%
        & 29347 & 1.406\%
        & 29802 & 2.979\%
        & 29385 & 1.538\% \\

        X-n134-k13
        & 10916
        & \textbf{11302} & \textbf{3.536\%}
        & 11581 & 6.092\%
        & 11522 & 5.551\%
        & 11353 & 4.003\%
        & 11418 & 4.599\% \\

        X-n139-k10
        & 13590
        & 14035 & 3.274\%
        & 13911 & 2.362\%
        & 14035 & 3.274\%
        & \textbf{13825} & \textbf{1.729\%}
        & 13892 & 2.222\% \\

        X-n143-k7
        & 15700
        & 16131 & 2.745\%
        & 16660 & 6.115\%
        & \textbf{16040} & \textbf{2.166\%}
        & 16125 & 2.707\%
        & 16101 & 2.554\% \\

        X-n148-k46
        & 43448
        & 49328 & 13.533\%
        & 50782 & 16.880\%
        & \textbf{45487} & \textbf{4.693\%}
        & 46758 & 7.618\%
        & 46347 & 6.672\% \\

        X-n153-k22
        & 21220
        & 32476 & 53.044\%
        & 26237 & 23.643\%
        & \textbf{23095} & \textbf{8.836\%}
        & 23793 & 12.125\%
        & 24044 & 13.308\% \\

        X-n157-k13
        & 16876
        & 17660 & 4.646\%
        & 17510 & 3.757\%
        & 17483 & 3.597\%
        & 17650 & 4.586\%
        & \textbf{17193} & \textbf{1.878\%} \\

        X-n162-k11
        & 14138
        & 14889 & 5.312\%
        & 14720 & 4.117\%
        & 14888 & 5.305\%
        & \textbf{14654} & \textbf{3.650\%}
        & 14793 & 4.633\% \\

        X-n167-k10
        & 20557
        & 21822 & 6.154\%
        & 21399 & 4.096\%
        & 21236 & 3.303\%
        & 21340 & 3.809\%
        & \textbf{21136} & \textbf{2.817\%} \\

        X-n172-k51
        & 45607
        & 49556 & 8.659\%
        & 56385 & 23.632\%
        & \textbf{47651} & \textbf{4.482\%}
        & 51292 & 12.465\%
        & 49428 & 8.378\% \\

        X-n176-k26
        & 47812
        & 54197 & 13.354\%
        & 57637 & 20.549\%
        & \textbf{53165} & \textbf{11.196\%}
        & 55520 & 16.121\%
        & 54023 & 12.990\% \\

        X-n181-k23
        & 25569
        & 37311 & 45.923\%
        & 26219 & 2.542\%
        & 26114 & 2.131\%
        & 26258 & 2.695\%
        & \textbf{26088} & \textbf{2.030\%} \\

        X-n186-k15
        & 24145
        & 25222 & 4.461\%
        & \textbf{25000} & \textbf{3.541\%}
        & 25605 & 6.047\%
        & 25182 & 4.295\%
        & 25135 & 4.100\% \\

        X-n190-k8
        & 16980
        & 18315 & 7.862\%
        & 18113 & 6.673\%
        & \textbf{18089} & \textbf{6.531\%}
        & 18327 & 7.933\%
        & 18649 & 9.829\% \\

        X-n195-k51
        & 44225
        & 49158 & 11.154\%
        & 54090 & 22.306\%
        & \textbf{47009} & \textbf{6.295\%}
        & 49984 & 13.022\%
        & 47962 & 8.450\% \\

        X-n200-k36
        & 58578
        & 64618 & 10.311\%
        & 61654 & 5.251\%
        & 61604 & 5.166\%
        & 61530 & 5.039\%
        & \textbf{61362} & \textbf{4.753\%} \\

        X-n209-k16
        & 30656
        & 32212 & 5.076\%
        & 32011 & 4.420\%
        & 32057 & 4.570\%
        & 32033 & 4.492\%
        & \textbf{31759} & \textbf{3.598\%} \\

        X-n219-k73
        & 117595
        & 133545 & 13.564\%
        & \textbf{119887} & \textbf{1.949\%}
        & 125029 & 6.322\%
        & 121046 & 2.935\%
        & 121064 & 2.950\% \\

        X-n228-k23
        & 25742
        & 48689 & 89.142\%
        & 33091 & 28.549\%
        & \textbf{28313} & \textbf{9.988\%}
        & 31054 & 20.636\%
        & 30247 & 17.501\% \\

        X-n237-k14
        & 27042
        & 29893 & 10.543\%
        & \textbf{28472} & \textbf{5.288\%}
        & 28924 & 6.960\%
        & 28550 & 5.577\%
        & 28580 & 5.687\% \\

        X-n247-k50
        & 37274
        & 56167 & 50.687\%
        & 45065 & 20.902\%
        & \textbf{41581} & \textbf{11.555\%}
        & 43673 & 17.167\%
        & 43552 & 16.843\% \\

        X-n251-k28
        & 38684
        & \textbf{40263} & \textbf{4.082\%}
        & 40614 & 4.989\%
        & 40538 & 4.793\%
        & 41022 & 6.044\%
        & 40525 & 4.759\% \\

        \midrule
        \multicolumn{2}{c}{Average Gap}
        & \multicolumn{2}{c}{16.629\%}
        & \multicolumn{2}{c}{9.820\%}
        & \multicolumn{2}{c}{\textbf{5.198\%}}
        & \multicolumn{2}{c}{6.884\%}
        & \multicolumn{2}{c}{5.894\%} \\
        \bottomrule
    \end{tabular}%
    }
\end{table*}

To evaluate zero-shot generalization on benchmark instances, we further tested the models on the Set-Solomon VRPTW benchmark without additional training. As reported in Table~\ref{tab:solomon_selected_results}, LaT-POMO-MTL reduced the average gap of POMO-MTL from 11.833\% to 10.723\%. LaT-MVMoE reduced the average gap of MVMoE from 10.336\% to 9.904\% and achieved the lowest average gap among the evaluated neural solvers. The LaT variants also obtained the lowest gaps on most benchmark instances. These results demonstrate that LaT improves out-of-distribution
generalization to Solomon VRPTW instances.

\begin{table*}[!t]
    \centering
    \caption{Zero-shot performance on the Set-Solomon VRPTW benchmark. All models are trained on problem size \(N=100\) following the experimental settings of \citep{zhou2024mvmoe}. The lowest objective value and its corresponding gap for each instance are highlighted in \textbf{bold}.}
    \label{tab:solomon_selected_results}
    \setlength{\tabcolsep}{3.2pt}
    \renewcommand{\arraystretch}{1.05}
    \resizebox{\textwidth}{!}{%
    \begin{tabular}{lrrrrrrrrrrr}
        \toprule
        \multirow{2}{*}{Instance}
        & \multirow{2}{*}{Opt.}
        & \multicolumn{2}{c}{POMO}
        & \multicolumn{2}{c}{POMO-MTL}
        & \multicolumn{2}{c}{LaT-POMO-MTL}
        & \multicolumn{2}{c}{MVMoE}
        & \multicolumn{2}{c}{LaT-MVMoE} \\
        \cmidrule(lr){3-4}
        \cmidrule(lr){5-6}
        \cmidrule(lr){7-8}
        \cmidrule(lr){9-10}
        \cmidrule(lr){11-12}
        & & Obj. & Gap
          & Obj. & Gap
          & Obj. & Gap
          & Obj. & Gap
          & Obj. & Gap \\
        \midrule

        R101
        & 1637.7
        & 1805.6 & 10.252\%
        & 1821.2 & 11.205\%
        & 1807.9 & 10.392\%
        & 1798.1 & 9.794\%
        & \textbf{1767.4} & \textbf{7.920\%} \\

        R102
        & 1466.6
        & 1556.7 & 6.143\%
        & 1596.0 & 8.823\%
        & 1574.7 & 7.370\%
        & 1572.0 & 7.187\%
        & \textbf{1544.5} & \textbf{5.313\%} \\

        R103
        & 1208.7
        & 1341.4 & 10.979\%
        & 1327.3 & 9.812\%
        & 1341.9 & 11.016\%
        & 1328.2 & 9.887\%
        & \textbf{1304.5} & \textbf{7.923\%} \\

        R104
        & 971.5
        & 1118.6 & 15.142\%
        & 1120.7 & 15.358\%
        & \textbf{1110.8} & \textbf{14.344\%}
        & 1124.8 & 15.780\%
        & 1127.5 & 16.058\% \\

        R105
        & 1355.3
        & 1506.4 & 11.149\%
        & 1514.6 & 11.754\%
        & \textbf{1440.0} & \textbf{6.251\%}
        & 1479.4 & 9.157\%
        & 1472.5 & 8.644\% \\

        R106
        & 1234.6
        & 1365.2 & 10.578\%
        & 1380.5 & 11.818\%
        & 1347.7 & 9.165\%
        & 1362.4 & 10.352\%
        & \textbf{1341.2} & \textbf{8.638\%} \\

        R107
        & 1064.6
        & 1214.2 & 14.052\%
        & 1209.3 & 13.592\%
        & 1208.0 & 13.472\%
        & 1182.1 & 11.037\%
        & \textbf{1180.2} & \textbf{10.859\%} \\

        R108
        & 932.1
        & 1058.9 & 13.604\%
        & 1061.8 & 13.915\%
        & 1055.3 & 13.217\%
        & \textbf{1023.2} & \textbf{9.774\%}
        & 1049.9 & 12.638\% \\

        R109
        & 1146.9
        & 1249.0 & 8.902\%
        & 1265.7 & 10.358\%
        & 1253.4 & 9.285\%
        & 1255.6 & 9.478\%
        & \textbf{1236.1} & \textbf{7.774\%} \\

        R110
        & 1068.0
        & 1180.4 & 10.524\%
        & 1171.4 & 9.682\%
        & \textbf{1165.0} & \textbf{9.082\%}
        & 1185.7 & 11.021\%
        & 1190.8 & 11.495\% \\

        R111
        & 1048.7
        & 1177.2 & 12.253\%
        & 1211.5 & 15.524\%
        & 1194.6 & 13.908\%
        & \textbf{1176.1} & \textbf{12.148\%}
        & 1183.5 & 12.854\% \\

        R112
        & 948.6
        & 1063.1 & 12.070\%
        & 1057.0 & 11.427\%
        & 1057.8 & 11.510\%
        & \textbf{1045.2} & \textbf{10.183\%}
        & 1052.0 & 10.896\% \\

        RC101
        & 1619.8
        & 2643.0 & 63.168\%
        & 1833.3 & 13.181\%
        & 1808.3 & 11.634\%
        & \textbf{1774.4} & \textbf{9.544\%}
        & 1821.5 & 12.453\% \\

        RC102
        & 1457.4
        & \textbf{1534.8} & \textbf{5.311\%}
        & 1546.1 & 6.086\%
        & 1565.5 & 7.421\%
        & 1544.5 & 5.976\%
        & 1553.2 & 6.574\% \\

        RC103
        & 1258.0
        & 1407.5 & 11.884\%
        & 1396.2 & 10.986\%
        & 1463.3 & 16.319\%
        & 1402.5 & 11.486\%
        & \textbf{1390.9} & \textbf{10.561\%} \\

        RC104
        & 1132.3
        & 1261.8 & 11.437\%
        & 1271.7 & 12.311\%
        & \textbf{1240.7} & \textbf{9.572\%}
        & 1265.4 & 11.755\%
        & 1243.9 & 9.854\% \\

        RC105
        & 1513.7
        & 1612.9 & 6.553\%
        & 1644.9 & 8.668\%
        & 1652.6 & 9.173\%
        & 1635.5 & 8.047\%
        & \textbf{1600.3} & \textbf{5.724\%} \\

        RC106
        & 1372.7
        & 1539.3 & 12.137\%
        & 1552.8 & 13.120\%
        & \textbf{1487.6} & \textbf{8.367\%}
        & 1505.0 & 9.638\%
        & 1493.6 & 8.811\% \\

        RC107
        & 1207.8
        & 1347.7 & 11.583\%
        & 1384.8 & 14.655\%
        & 1333.5 & 10.411\%
        & 1351.6 & 11.906\%
        & \textbf{1329.3} & \textbf{10.057\%} \\

        RC108
        & 1114.2
        & 1305.5 & 17.169\%
        & 1274.4 & 14.378\%
        & \textbf{1254.1} & \textbf{12.557\%}
        & 1254.2 & 12.565\%
        & 1259.4 & 13.030\% \\

        \midrule
        \multicolumn{2}{c}{Average Gap}
        & \multicolumn{2}{c}{13.745\%}
        & \multicolumn{2}{c}{11.833\%}
        & \multicolumn{2}{c}{10.723\%}
        & \multicolumn{2}{c}{10.336\%}
        & \multicolumn{2}{c}{\textbf{9.904\%}} \\
        \bottomrule
    \end{tabular}%
    }
\end{table*}

\section{Complexity Analysis}
\label{app:complexity-analysis}

Table~\ref{tab:complexity-analysis} compares the training cost of LaT with
that of each backbone. All measurements are obtained for 50-node VRPs on a
single NVIDIA RTX A6000 GPU. Each model is jointly trained on the six training
variants and evaluated on all \(16\) variants. Training time covers the complete
\(5{,}000\)-epoch training process, including the predefined LLM calls. Peak
memory is the maximum GPU memory recorded during joint training. The parameter
count includes all encoder and decoder parameters, including the LaT side
branches. The pretrained LLM is accessed through an API, so its parameters are not
included in the deployed neural solver.

Across the six backbones, LaT increases the total training time by only
\(0.08\) to \(0.24\) hours. The corresponding relative increase ranges from
\(0.27\%\) to \(1.49\%\), with an average of \(0.71\%\). Peak GPU memory
increases by only \(8\) to \(22\) MiB, remaining below \(0.29\%\) for every
backbone. The encoder side branches add approximately \(0.20\) to \(0.21\)
million parameters. These results show that the predefined LLM calls and
retained side branches introduce little additional training cost.


\begin{table}[htbp]
    \centering
    \caption{Training cost and model size for 50-node multi-task VRPs.}
    \label{tab:complexity-analysis}

    \footnotesize
    \setlength{\tabcolsep}{3.6pt}
    \renewcommand{\arraystretch}{1.12}

    \begin{tabular}{@{}lccc@{}}
        \toprule
        \multicolumn{1}{l}{Method}
        & Peak Memory (MiB) $\downarrow$
        & Params (M) $\downarrow$
        & Training Time (h) $\downarrow$ \\
        \midrule

        POMO-MTL
        & 6,780
        & 1.25
        & 16.11 \\

        LaT-POMO-MTL
        & 6,792
        & 1.46
        & 16.35 \\

        MVMoE-L
        & 9,148
        & 3.70
        & 30.03 \\

        LaT-MVMoE-L
        & 9,157
        & 3.91
        & 30.11 \\

        MVMoE
        & 9,584
        & 3.68
        & 36.05 \\

        LaT-MVMoE
        & 9,602
        & 3.89
        & 36.15 \\

        ReLD
        & 7,672
        & 1.39
        & 20.69 \\

        LaT-ReLD
        & 7,694
        & 1.59
        & 20.90 \\

        CaDA
        & 9,750
        & 3.42
        & 18.03 \\

        LaT-CaDA
        & 9,762
        & 3.63
        & 18.18 \\

        CCL
        & 19,788
        & 1.57
        & 41.67 \\

        LaT-CCL
        & 19,796
        & 1.77
        & 41.84 \\

        \bottomrule
    \end{tabular}
\end{table}

\section{Ablation Study}

\subsection{Comparison with an MLP Guidance Generator}
\label{app:mlp-guidance-comparison}

To examine the contribution of the pretrained LLM, we construct an MLP-based
guidance baseline by replacing only the LLM guidance vector generator with a
learnable neural network. The POMO-MTL backbone, encoder side branches,
training configuration, guidance update interval, and evaluation protocol
remain unchanged.

At each control round \(r\), the MLP baseline converts the available numerical
training state into a fixed-dimensional input vector
\(\boldsymbol{\chi}_{r}\). This vector contains the constraint indicators,
the initial guidance, the guidance used during the preceding training
interval, validation objectives, relative reference gaps, gap changes, and
recent objective and gap histories of the \(16\) VRP variants. The MLP first
applies layer normalization and then uses two fully connected hidden layers,
each with \(64\) units and ReLU activation. A final linear layer produces five
logits, which are mapped to the predefined guidance range through a sigmoid
function:
\begin{equation}
\boldsymbol{\gamma}^{\mathrm{MLP}}_{r}
=
\gamma_{\min}^{\mathrm{MLP}}\mathbf{1}_5
+
\left(
\gamma_{\max}^{\mathrm{MLP}}
-
\gamma_{\min}^{\mathrm{MLP}}
\right)
\sigma
\left(
\operatorname{MLP}_{\phi}
\left(
\operatorname{LN}(\boldsymbol{\chi}_{r})
\right)
\right),
\label{eq:mlp-guidance}
\end{equation}
where \(\boldsymbol{\gamma}^{\mathrm{MLP}}_{r}\in\mathbb{R}^{5}\) contains
the guidance values for C, O, B, L, and TW, and \(\phi\) denotes the parameters
of the MLP guidance generator. We set
\(\gamma_{\min}^{\mathrm{MLP}}=0.5\) and
\(\gamma_{\max}^{\mathrm{MLP}}=2.0\). The output layer is initialized so that
the initial MLP output is \(\mathbf{1}_5\), consistent with the initial
guidance used by LaT.

The MLP parameters are updated once per control round. In the first training
batch of the subsequent interval, the guidance vector remains connected to the
MLP parameters, allowing the original policy-gradient loss to update
\(\phi\) through the encoder side branches. No auxiliary loss or explicit
guidance labels are used. After this update, the guidance vector is recomputed
using the updated MLP, detached from the computational graph, and fixed until
the next control round. The solver parameters \(\theta\) and the coefficient
\(\alpha\) continue to be updated in every training batch.

\begin{figure*}[htbp]
  \centering
  \includegraphics[width=0.92\textwidth]{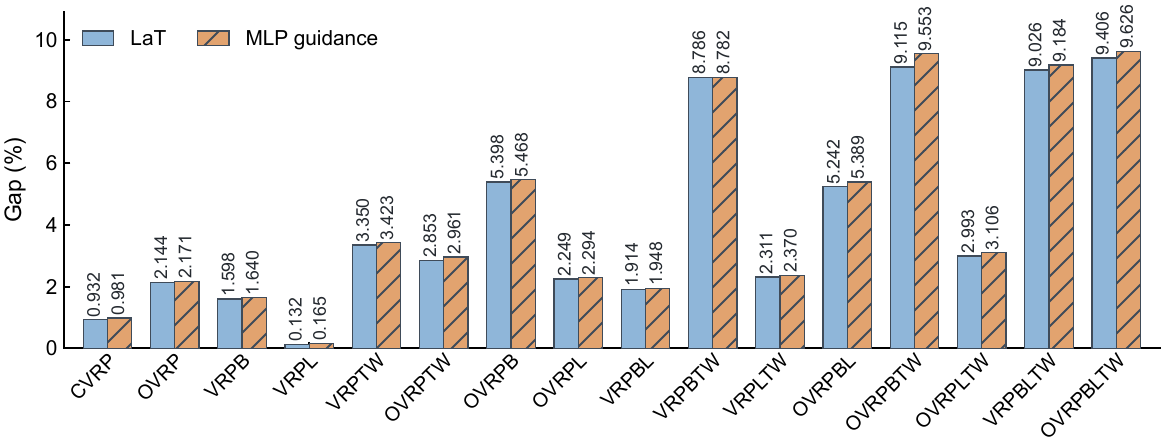}
  \caption{Comparison of relative reference gaps between the LLM-based LaT
  model and the MLP guidance baseline on 50-node multi-task VRPs. Lower values
  indicate better performance.}
  \label{fig:llm-vs-mlp-gap}
\end{figure*}

As shown in Fig.~\ref{fig:llm-vs-mlp-gap}, LaT achieved an average gap of
4.216\%, compared with 4.316\% for the MLP guidance baseline, and obtained a
lower gap on 15 of the 16 VRP variants. The average gap was reduced by
0.100\%, with more evident improvements on multi-constraint variants including
OVRPBTW, VRPBLTW, and OVRPBLTW. The MLP guidance generator introduces
additional trainable parameters on top of the same POMO-MTL backbone and
encoder side branches. However, its average gap remains higher than that of
LaT. Therefore, the improvement of LaT cannot be explained by the increase in
trainable parameters. Together with the component ablation in
Fig.~\ref{fig:component-ablation}, this result shows that the LLM-generated
guidance vector contributes to solution quality.

\subsection{Ablation on Guidance Injection Position}
\label{app:injection-ablation}

We conduct this ablation to examine where the LaT guidance vector should be injected
into the neural solver. All variants use the same POMO-MTL backbone, training
configuration, and evaluation protocol. Only the injection position of the side
branch is changed. LaT Input injects the guidance vector once after the initial node
embedding and before the first encoder layer. LaT EncLast injects it once after
the final encoder layer. LaT DecCtx injects it into the decoder context after
multi-head attention and before final node scoring. LaT EncAll is the proposed
design, in which the guidance vector is injected after every encoder layer.

As shown in Table~\ref{tab:lat-injection-ablation}, LaT EncAll achieves the lowest average score of \(10.090\) and the lowest
average gap of \(4.216\%\). LaT Input, LaT DecCtx, and LaT EncLast obtain
average gaps of \(4.261\%\), \(4.276\%\), and \(4.462\%\), respectively.
These results support injecting the side branch after every encoder layer in
the evaluated setting.

\begin{table}[htbp]
\centering
\caption{Ablation on the injection position of LaT guidance on 50-node multi-task VRPs.}
\label{tab:lat-injection-ablation}
\begin{tabular}{lcc}
\toprule
Variant & Avg. Score $\downarrow$ & Avg. Gap $\downarrow$ \\
\midrule
LaT Input   & 10.093 & 4.261\% \\
LaT EncLast & 10.111 & 4.462\% \\
LaT DecCtx  & 10.094 & 4.276\% \\
LaT EncAll & \textbf{10.090} & \textbf{4.216\%} \\
\bottomrule
\end{tabular}
\end{table}
\end{document}

%% file: math_commands.tex
\usepackage{amsmath,amsfonts,bm}

\def\eqref#1{equation~\ref{#1}}

\def\1{\bm{1}}

\DeclareMathAlphabet{\mathsfit}{\encodingdefault}{\sfdefault}{m}{sl}
\SetMathAlphabet{\mathsfit}{bold}{\encodingdefault}{\sfdefault}{bx}{n}

